\documentclass{article}

\usepackage{arxiv}

\usepackage[utf8]{inputenc}
\usepackage[T1]{fontenc}
\usepackage{hyperref}
\usepackage{url}
\usepackage{booktabs}
\usepackage{amsmath,amssymb,amsfonts}
\usepackage{nicefrac}
\usepackage{microtype}
\usepackage{graphicx}
\usepackage[numbers,sort&compress]{natbib}
\usepackage{doi}
\usepackage{array}
\usepackage{xcolor}
\usepackage{enumitem}
\usepackage{caption}
\usepackage{algorithm}
\usepackage{algpseudocode}
\usepackage{multirow}
\usepackage{tabularx}
\usepackage[section]{placeins}
\usepackage{tikz}
\usetikzlibrary{positioning,arrows.meta,fit,backgrounds,calc}

\graphicspath{{./}}
\newcommand{\cls}[1]{\texttt{#1}}
\newcommand{\nb}[1]{Notebook~#1}
\newcommand{\pp}{\,pp}

\newcommand{\schoolurl}{https://school.flta-conference.org/}
\newcommand{\indigmaurl}{https://indigma.eu/}

\newcommand{\repourl}{https://github.com/Indigma-Innovations/fl-summer-school}

\title{Federated Art Chase: Privacy Leakage and Data Poisoning in Federated Fine-Tuning}

\author{ \href{https://orcid.org/0000-0003-3915-9628}{\includegraphics[scale=0.06]{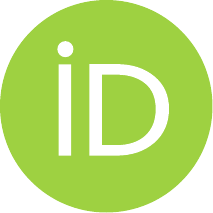}\hspace{1mm}Vasileios Perifanis} \\
	\href{\indigmaurl}{Indigma Innovations}\\
	Greece \\
	\texttt{vasilis@indigma.eu} \\
	\And
	\href{https://orcid.org/0000-0001-9370-5023}{\includegraphics[scale=0.06]{orcid.pdf}\hspace{1mm}Nikolaos Pavlidis} \\
	\href{\indigmaurl}{Indigma Innovations}\\
	Greece \\
	\texttt{nikos@indigma.eu} \\
}

\renewcommand{\headeright}{Indigma}
\renewcommand{\undertitle}{\href{\schoolurl}{4th Federated Machine Learning International Summer School}}
\renewcommand{\shorttitle}{Federated Art Chase}

\hypersetup{
pdftitle={Federated Art Chase: Privacy Leakage and Data Poisoning in Federated Fine-Tuning},
pdfauthor={Vasileios Perifanis, Nikolaos Pavlidis},
pdfkeywords={Federated Learning, Transfer Learning, Gradient Inversion, Label Leakage, Data Poisoning, Backdoor Attack, Byzantine-Robust Aggregation, FLTrust},
colorlinks=true,
linkcolor=blue!50!black,
citecolor=blue!50!black,
urlcolor=blue!50!black,
}

\begin{document}
\maketitle

\begin{abstract}
Federated learning (FL) is increasingly used to adapt a pretrained, centrally fine-tuned model to data that remain on user devices, beyond training a model from scratch in a distributed manner. In this report we study the privacy and integrity of such an approach end to end, using a single reproducible pipeline prepared for the 4th Federated Machine Learning International Summer School. An ImageNet-pretrained MobileNetV4-Conv-Small model is first fine-tuned centrally on ten classes of the Quick, Draw! dataset, which raises test accuracy from 6.00\% with an untrained classifier head to 97.50\%. The resulting checkpoint initializes head-only FedAvg over ten non-IID clients. After 50 rounds, client training accuracy rises from 90.34\% to 91.49\%, while accuracy on a global evaluation set drawn from the Stage~1 data is preserved (97.66\% before and 97.64\% after). On this pipeline we evaluate four adversarial scenarios. First, from a single one-step client update, an honest-but-curious server recovers the complete private label set and reconstructs class-plausible sketches, although pixel fidelity remains low (mean SSIM 0.130) and the reconstructions follow class prototypes rather than the private drawings. Second, from a one-round FedAvg aggregate, label-count recovery falls from 87.5\% with one client to 75.0\% with five and ten clients, which is below the 80.0\% of an uninformed uniform guess. Third, under FedAvg, a backdoor injected by 30\%, 50\%, and 100\% of the clients reaches an attack success rate (ASR) of 21.9\%, 31.3\%, and 54.7\%, respectively, while reducing the final clean accuracy by at most 0.8 percentage points. Label flipping, in contrast, becomes effective only when the attacker relabels most of the source-class data. Fourth, FLTrust with a 1{,}000-sample root set retained from centralized fine-tuning lowers the backdoor ASR to 1.8\% and 4.0\% at 30\% and 50\% malicious clients and the label-flipping ASR from 12.6\% to 0.8\%, without a measurable loss of clean accuracy. It provides no protection, however, when every client is malicious. This report closes with a live session in which participants draw, fine-tune, and poison the Stage~1 model through Indigma's FedMaestro orchestrator. Our code and executed notebooks are available at \url{\repourl} and run on Google Colab without specialized hardware.
\end{abstract}


\newpage
\tableofcontents
\newpage

\section{Introduction}
\label{sec:intro}

Federated learning (FL) trains a shared model while the raw training data remain with the parties that hold them. A server receives model updates and combines them using a federated aggregation algorithm such as FedAvg \citep{mcmahan2017fedavg,kairouz2021advances}. Tutorials and benchmark studies of FL frequently start from a randomly initialized network that is trained entirely through federated rounds. Such a setting is convenient for controlled comparisons, but it differs from how many vision and language systems are built in practice. A typical development path starts from a backbone pretrained on a large public corpus, adapts it centrally to the target task with representative data available, and only then uses FL to adapt the resulting model to data that cannot be centralized. Pretrained initialization is known to improve the accuracy and stability of federated optimization \citep{nguyen2023where,chen2023importance}, while adapting only a small task-specific part of the network is a common way to personalize a shared representation at a low cost \citep{arivazhagan2019personalization,collins2021fedrep}.

This deployment pattern changes the security question. The model that enters the federated stage is already accurate, the backbone may be frozen, and each client may update only a few thousand parameters. Two families of risk nevertheless remain. On the one hand, model updates are deterministic functions of the private data that produced them, and an honest-but-curious server can therefore try to infer labels or reconstruct inputs from them \citep{melis2019exploiting,zhu2019dlg,geiping2020inverting}. On the other hand, malicious clients can poison their local data or updates to degrade the global model or to implant a backdoor \citep{tolpegin2020datapoison,bagdasaryan2020backdoor,bhagoji2019adversarial}, and robust aggregation rules have been proposed to limit their influence \citep{blanchard2017krum,yin2018byzantine,cao2021fltrust}. Most of these results were obtained for full-model training, often from scratch. How strongly they apply when FL is the last, head-only stage of a pretrained pipeline is less documented, and this is the question we study in this report.

The report accompanies a hands-on tutorial at the 4th Federated Machine Learning International Summer School\footnote{\url{\schoolurl}}, held in Paris, France, in October 2026. All experiments are implemented as six connected notebooks that share one task, one model, and one set of persisted artifacts. Every number in this report can be traced to a specific executed notebook. The notebooks run on Google Colab, i.e., participants can reproduce every experiment in a browser without local installation or high-end hardware. The pipeline fine-tunes an ImageNet-pretrained MobileNetV4-Conv-Small model \citep{qin2024mobilenetv4} on ten classes of the Quick, Draw! dataset \citep{quickdraw,ha2018sketchrnn}, uses the resulting checkpoint to initialize head-only FedAvg over ten non-IID clients, and then attacks and defends the federated stage. The tutorial closes with a live session on FedMaestro, the federated learning orchestrator developed by Indigma\footnote{\url{\indigmaurl}}.

The main contributions of this report are summarized as follows:
\begin{itemize}[leftmargin=1.5em,itemsep=2pt]
    \item We describe a deployment-oriented FL pipeline that consists of pretraining, centralized fine-tuning, and federated fine-tuning, with explicit data provenance. The data used by each stage and by each adversary are drawn from documented index ranges, and the private data of every experiment are disjoint from the Stage~1 training data and from the public data available to the adversary. We also document that the global evaluation set of the federated notebooks is not the held-out Stage~1 test set, and we quantify the overlap.
    \item We show that, under one-step local SGD, the server recovers the exact classifier-head gradient both from a single client update and from a FedAvg aggregation. From a single update the server recovers the full private label set, but only within a narrow range of a heuristic weight. Pixel-space inversion yields class-plausible sketches of low fidelity, and under aggregation, label-count recovery falls below an uninformed baseline.
    \item We evaluate label flipping and a pixel-trigger backdoor for five malicious-client fractions under FedAvg. We show that the attack strength is explained by the amount of poisoned data that reaches the aggregate rather than by the nominal malicious fraction.
    \item We evaluate FLTrust with a root set retained from centralized fine-tuning. Using per-round trust diagnostics, we quantify how much aggregation weight FLTrust grants to malicious clients compared with FedAvg.
    \item We describe a live session in which the Stage~1 model is deployed in Indigma's FedMaestro orchestrator, so that participants can observe inference, federated fine-tuning, and participant-driven poisoning on a running system.
\end{itemize}

The rest of this report is organized as follows. Section~\ref{sec:background} reviews related work on FL from pretrained models, privacy leakage from model updates, and data poisoning. Section~\ref{sec:pipeline} describes the experimental pipeline, while Sections~\ref{sec:stage1} and~\ref{sec:stage2} present the centralized and the clean federated fine-tuning stages. Sections~\ref{sec:privacy}, \ref{sec:poisoning}, and~\ref{sec:defense} evaluate the privacy attacks, the poisoning attacks, and the FLTrust defense, respectively. Section~\ref{sec:resources} reports the computational footprint of every experiment, and Section~\ref{sec:fedmaestro-live} describes the live session. Section~\ref{sec:discussion} discusses lessons learned and limitations. Finally, Section~\ref{sec:conclusion} concludes the report and outlines directions for future work.

\section{Related Work}
\label{sec:background}

\paragraph{Federated learning from pretrained models.}
In FedAvg \citep{mcmahan2017fedavg}, a server broadcasts the current global model to a set of clients, and each client performs a few local optimization steps on its private data. The server then updates the global model by the average of the returned models, weighted by the number of local training samples. Statistical heterogeneity across clients (non-IID data) slows convergence and is commonly simulated by drawing per-client class proportions from a Dirichlet distribution with a small concentration parameter \citep{hsu2019dirichlet}. Starting federated optimization from a pretrained model reduces the harm caused by heterogeneity and closes much of the gap to centralized training \citep{nguyen2023where,chen2023importance}. Pretrained ImageNet \citep{deng2009imagenet} features transfer well to new visual tasks \citep{yosinski2014transferable,kornblith2019better}, and splitting a network into a shared representation and a lightweight, locally adapted head is a standard approach to federated personalization \citep{arivazhagan2019personalization,collins2021fedrep}. Our pipeline combines both ideas, i.e., the backbone is adapted centrally and then frozen, while FL updates the final linear classifier.

\paragraph{Privacy leakage from model updates.}
Model updates leak information about the data that produced them, including properties that are unrelated to the learning task \citep{melis2019exploiting}. Deep Leakage from Gradients (DLG) reconstructs training images and labels by optimizing dummy inputs until their gradient matches an observed gradient \citep{zhu2019dlg}. iDLG showed that, for a single sample trained with cross-entropy, the label can be read directly from the sign of the last-layer gradient \citep{zhao2020idlg}, and later work extended label inference to batches and to aggregated updates \citep{yin2021gradinversion,wainakh2022label}. Geiping et al.\ replaced the Euclidean matching loss by a cosine loss with a total-variation prior \citep{rudin1992tv} and reconstructed ImageNet-scale images from realistic updates \citep{geiping2020inverting}. GradInversion further added batch-level label recovery and image priors \citep{yin2021gradinversion}. Huang et al.\ showed that reported attack strength depends heavily on assumptions such as known labels, known batch-normalization statistics, and small batches \citep{huang2021evaluating}, which motivates the explicit statement of assumptions in Section~\ref{sec:privacy}. Secure aggregation \citep{bonawitz2017secagg} hides individual updates from the server, and differentially private training \citep{abadi2016dpsgd} bounds what any update can reveal. However, neither privacy-preserving technique is applied in this report.

\paragraph{Data poisoning and robust aggregation.}
In data poisoning, malicious participants train on manipulated data by following the federated protocol. Label flipping relabels examples of a source class as a target class and degrades the source class in proportion to the number of poisoned participants \citep{tolpegin2020datapoison}. Backdoor attacks implant a trigger that redirects predictions to an attacker-chosen class while leaving clean accuracy largely intact \citep{gu2017badnets}. In FL, backdoors can be injected through model replacement \citep{bagdasaryan2020backdoor} or boosted malicious updates \citep{bhagoji2019adversarial}, and their persistence depends on the fraction of attackers and on norm constraints \citep{sun2019backdoor,wang2020tails}. Byzantine-robust aggregation rules limit the influence of outlying updates. Krum selects the update closest to its neighbours \citep{blanchard2017krum}, the coordinate-wise median and trimmed mean discard extreme values \citep{yin2018byzantine}, and Bulyan combines both ideas \citep{elmhamdi2018bulyan}. Such rules can be circumvented by optimized attacks \citep{fang2020local,shejwalkar2021manipulating}, and their assumptions are strained by non-IID honest clients. FLTrust bootstraps trust from a small clean root dataset held by the server. Each client update is scored by its cosine similarity with a server update computed on the root set, and accepted updates are rescaled to the norm of the server update \citep{cao2021fltrust}. In our pipeline, a retained root set is a natural extension of the centralized stage, which makes FLTrust a well-matched defense.

\section{Experimental Pipeline}
\label{sec:pipeline}

\subsection{Overview}
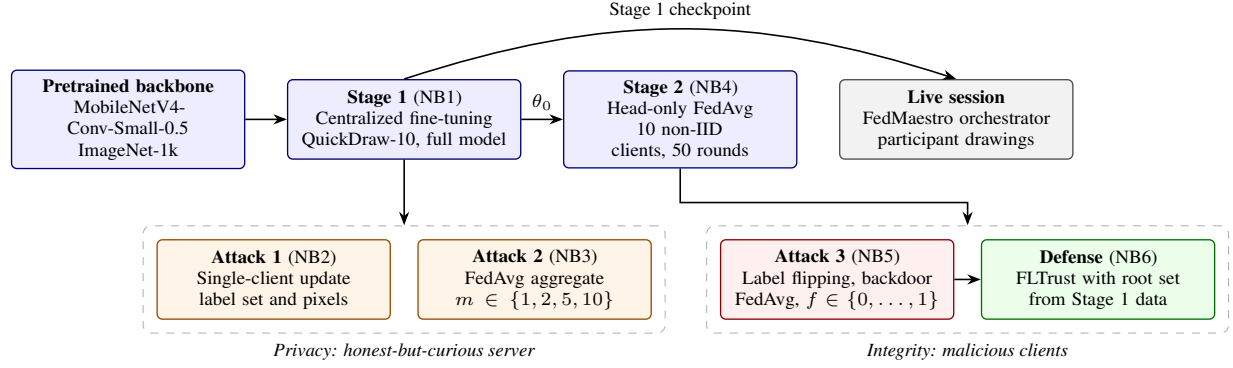
\begin{figure}[!tbp]
\centering
\begin{tikzpicture}[
    node distance=0.45cm and 0.55cm,
    box/.style={draw, rounded corners=2pt, minimum height=1.05cm, text width=2.85cm, align=center, font=\scriptsize, line width=0.5pt},
    stage/.style={box, fill=blue!7, draw=blue!45!black},
    priv/.style={box, fill=orange!9, draw=orange!60!black},
    integ/.style={box, fill=red!6, draw=red!55!black},
    defn/.style={box, fill=green!8, draw=green!45!black},
    live/.style={box, fill=gray!10, draw=gray!60!black},
    arrow/.style={-{Stealth[length=1.8mm]}, line width=0.6pt},
    grp/.style={draw=gray!60, dashed, rounded corners=3pt, inner sep=5pt}
]
\node[stage] (pre) {\textbf{Pretrained backbone}\\MobileNetV4-Conv-Small-0.5\\ImageNet-1k};
\node[stage, right=of pre] (s1) {\textbf{Stage 1} (NB1)\\Centralized fine-tuning\\QuickDraw-10, full model};
\node[stage, right=of s1] (s2) {\textbf{Stage 2} (NB4)\\Head-only FedAvg\\10 non-IID clients, 50 rounds};
\node[live, right=of s2] (live) {\textbf{Live session}\\FedMaestro orchestrator\\participant drawings};

\node[priv, below=1.05cm of s1, anchor=north east, xshift=-0.175cm] (a1) {\textbf{Attack 1} (NB2)\\Single-client update\\label set and pixels};
\node[priv, right=0.35cm of a1] (a2) {\textbf{Attack 2} (NB3)\\FedAvg aggregate\\$m\in\{1,2,5,10\}$};
\node[integ, right=0.9cm of a2] (a3) {\textbf{Attack 3} (NB5)\\Label flipping, backdoor\\FedAvg, $f\in\{0,\dots,1\}$};
\node[defn, right=0.35cm of a3] (d4) {\textbf{Defense} (NB6)\\FLTrust with root set\\from Stage 1 data};

\begin{scope}[on background layer]
\node[grp, fit=(a1)(a2), label={[font=\scriptsize\itshape]below:Privacy: honest-but-curious server}] (gp) {};
\node[grp, fit=(a3)(d4), label={[font=\scriptsize\itshape]below:Integrity: malicious clients}] (gi) {};
\end{scope}

\draw[arrow] (pre) -- (s1);
\draw[arrow] (s1) -- node[above, font=\scriptsize]{$\theta_0$} (s2);
\draw[arrow] (s1.north) to[bend left=18] node[above, font=\scriptsize, pos=0.5]{Stage 1 checkpoint} (live.north);
\draw[arrow] (s1.south) -- (s1.south |- gp.north);
\draw[arrow] (s2.south) -- ++(0,-0.45) -| (gi.north);
\draw[arrow] (a3) -- (d4);
\end{tikzpicture}
\caption{Overview of the pipeline. Blue boxes denote model development stages, orange boxes privacy attacks by an honest-but-curious server, red and green boxes poisoning by malicious clients and the FLTrust defense, and the gray box the live session. NB$k$ denotes the executed notebook that implements each component. Attacks~1 and~2 start from the Stage~1 checkpoint, whereas Attack~3 and the defense reuse the exact client population of Stage~2.}
\label{fig:pipeline}
\end{figure}

Figure~\ref{fig:pipeline} summarizes the overall pipeline. Stage~1 (\nb{1}) fine-tunes the complete pretrained network centrally. The produced checkpoint is the common starting point of every later experiment, namely the single-client and aggregate privacy attacks (\nb{2} and~\nb{3}), the clean federated baseline (\nb{4}), the poisoning experiments under FedAvg (\nb{5}) and FLTrust (\nb{6}), and the live FedMaestro session. Table~\ref{tab:scenarios} lists the adversary, its capabilities, and the reported metrics for each experiment. Unless stated otherwise, all federated experiments train only the classifier head and all clients participate in every round. 

\begin{table}[!tbp]
\centering
\caption{Experiments, adversaries, and reported metrics. $\theta_t$ denotes the global classifier head at round $t$, $\Delta$ a model update, $n_k$ the sample count of client $k$, and $\eta$ the local learning rate.}
\label{tab:scenarios}
\small
\begin{tabularx}{\linewidth}{@{}l l X X@{}}
\toprule
Experiment & Adversary & Capabilities and assumptions & Reported metrics \\
\midrule
Stage 1 (NB1) & none & centralized training on QuickDraw-10 & loss, accuracy \\
Stage 2 (NB4) & none & 10 clients, head-only FedAvg & loss, accuracy (client and global) \\
Attack 1 (NB2) & honest-but-curious server & observes $\theta_t$, one client delta $\Delta_k$, $n_k$, $\eta$; public QuickDraw data & label-set accuracy; MSE, PSNR, SSIM \\
Attack 2 (NB3) & honest-but-curious server & observes $\theta_t$, $\theta_{t+1}$, $m$, $N=\sum_k n_k$, $\eta$; public QuickDraw data & label-count accuracy; MSE, PSNR, SSIM \\
Attack 3 (NB5) & malicious clients (fraction $f$) & poison local training data; follow the protocol otherwise & clean accuracy, ASR \\
Defense (NB6) & malicious clients (fraction $f$) & as Attack 3; server holds a clean root set & clean accuracy, ASR, trust scores \\
\bottomrule
\end{tabularx}
\end{table}

\subsection{Task and data}
\label{sec:data}
\begin{figure}[!tbp]
\centering
\includegraphics[width=\linewidth]{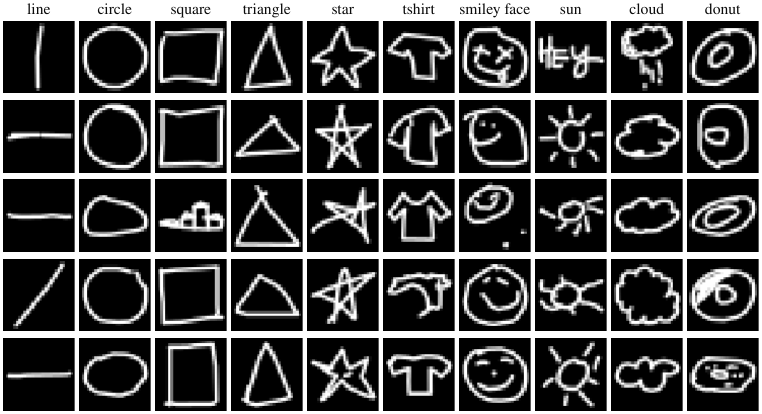}
\caption{Five randomly drawn Stage~1 examples per class ($28\times28$ bitmaps, white strokes on a black background).}
\label{fig:samples}
\end{figure}

The task is ten-class sketch recognition on the Quick, Draw! dataset \citep{quickdraw,ha2018sketchrnn}, which contains crowd-sourced drawings collected through an online game. We use the preprocessed $28\times28$ grayscale bitmaps of the classes \cls{line}, \cls{circle}, \cls{square}, \cls{triangle}, \cls{star}, \cls{tshirt}, \cls{smiley\_face}, \cls{sun}, \cls{cloud}, and \cls{donut} (Figure~\ref{fig:samples}). The ten class files contain between 120{,}265 (\cls{cloud}) and 143{,}549 (\cls{line}) bitmaps each. The classes combine simple geometric primitives with composite drawings and include visually related pairs such as \cls{circle}/\cls{donut} and \cls{sun}/\cls{star}, which become important during the execution of attacks.

Since every experiment draws from the same class files, we separate data by index range wherever disjointness matters, as summarized in Table~\ref{tab:provenance}. The Stage~1 split uses the first 5{,}000 bitmaps of each class. Federated clients receive bitmaps with index at least 10{,}000, and the private samples of the privacy attacks are drawn from indices of at least 20{,}000. The public prior knowledge of the attacker is drawn from indices below 20{,}000 and therefore never contains a private sample of the privacy experiments. The FLTrust root set is sampled from the Stage~1 data, which a provider would plausibly retain after centralized fine-tuning.

The federated notebooks (\nb{4}--\nb{6}) evaluate the global accuracy on 5{,}000 images obtained by re-running the stratified 80/10/10 split of the first 5{,}000 bitmaps per class. They use seed 123 instead of the seed 42 used in \nb{1}, so this \emph{global evaluation set} is a different random partition from the Stage~1 test set. Reproducing both splits shows that 3{,}990 of its images (79.8\%) belong to the Stage~1 training split, 479 (9.6\%) to the Stage~1 validation split, and only 531 (10.6\%) to the Stage~1 test split. The global evaluation set is disjoint from all federated client data and identical across all federated runs, and remains valid for comparing federated configurations with one another.

\begin{table}[!tbp]
\centering
\caption{Data provenance. Indices refer to positions within each QuickDraw class file. ``Hard'' selection retains the candidates with the highest cross-entropy loss under the Stage~1 model (Sections~\ref{sec:stage2} and~\ref{sec:privacy}).}
\label{tab:provenance}
\small
\begin{tabularx}{\linewidth}{@{}l l l X@{}}
\toprule
Use & Indices & Size & Selection \\
\midrule
Stage 1 train / validation / test (NB1) & $0$--$4{,}999$ & 40{,}000 / 5{,}000 / 5{,}000 & stratified 80/10/10 split per class (seed 42) \\
Federated clients (NB4--NB6) & $\geq 10{,}000$ & 13{,}675 (12{,}310 train, 1{,}365 val.) & hard selection from $\geq 5{,}000$ random candidates per class \\
Attack 1 private batch (NB2) & $\geq 20{,}000$ & 5 & hardest of 128 random candidates per class \\
Attack 2 private pool (NB3) & $\geq 20{,}000$ & 56 (10 clients) & hardest of 128 random candidates per class and client \\
Attacker public priors (NB2, NB3) & $< 20{,}000$ & 96 per class (prototypes), 1 per canvas & uniform random \\
Global evaluation set (NB4--NB6) & $0$--$4{,}999$ & 5{,}000 & test part of the 80/10/10 split rebuilt with seed 123 (79.8\% overlap with the Stage~1 training split) \\
FLTrust root set (NB6) & $0$--$4{,}999$ & 1{,}000 (100 per class) & uniform random from the training part of the seed-123 split \\
\bottomrule
\end{tabularx}
\end{table}

\subsection{Model and preprocessing}
\label{sec:model}
We use \texttt{mobilenetv4\_conv\_small\_050.e3000\_r224\_in1k}, the 0.5-width variant of the MobileNetV4-Small \citep{qin2024mobilenetv4} pretrained on ImageNet-1k obtained via \texttt{timm} \citep{rw2019timm}. Its ImageNet head is replaced by a randomly initialized linear layer with $C=10$ outputs, which results in 970{,}762 parameters. The backbone $f_\phi$ maps an input to a $d=1280$-dimensional embedding $h=f_\phi(x)$, and the head computes logits $z = Wh+b$ with $W\in\mathbb{R}^{C\times d}$ and $b\in\mathbb{R}^{C}$. The head has $C(d+1)=12{,}810$ parameters, i.e., 1.32\% of the network. In 32-bit floating point the parameters of the full model occupy 3.70\,MiB and those of the head 0.049\,MiB, so a head-only update is about $1/76$ the size of a full-model update.

Each bitmap is replicated to three channels and processed with the data configuration of the pretrained model, namely resizing to $256$ pixels, center cropping to $224\times224$, and normalization with mean and standard deviation $0.5$ per channel. During Stage~1 training, the pretrained configuration additionally applies random resized cropping (scale $0.08$--$1$), horizontal flipping with probability $0.5$, and color jitter of strength $0.4$. Stage~1 resizes with bicubic interpolation, as specified by the model configuration. \nb{2}--\nb{6}, in contrast, rebuild the evaluation transform from the stored mean, standard deviation, and input size and therefore use the \texttt{timm} default, i.e., bilinear interpolation. Together with the different evaluation split described in Section~\ref{sec:data}, this explains why the same checkpoint scores 97.50\% on the Stage~1 test set in \nb{1} and 97.66\% on the global evaluation set in \nb{4}--\nb{6}. The two numbers are not comparable, and we report each number with the preprocessing and data of the notebook that produced it.

\subsection{Notation and head gradients}
\label{sec:notation}
All federated experiments update only $\theta=(W,b)$. For a batch $\{(x_i,y_i)\}_{i=1}^{n}$ with embeddings $h_i$ and softmax probabilities $p_i=\mathrm{softmax}(Wh_i+b)$, the gradient of the mean cross-entropy loss with respect to the head is
\begin{equation}
    g_b = \frac{1}{n}\sum_{i=1}^{n}\left(p_i - e_{y_i}\right),
    \qquad
    g_W = \frac{1}{n}\sum_{i=1}^{n}\left(p_i - e_{y_i}\right)h_i^{\top},
    \label{eq:head-grad}
\end{equation}
where $e_{y}$ is the one-hot vector of label $y$. Row $c$ of $g_W$, denoted $g_W^{(c)}$, and entry $c$ of $g_b$, denoted $g_b^{(c)}$, correspond to output $c$ and aggregate all samples with weights $p_i^{(c)}-\mathbf{1}[y_i=c]$. The weight is positive for every $c\neq y_i$ and negative for $c=y_i$, since $0<p_i^{(c)}<1$ for finite logits\footnote{In floating point, $p_i^{(y_i)}$ can round to exactly 1 for very confident predictions, in which case the true-class weight is zero rather than negative.}. Hence the entry $g_b^{(c)}$ can be negative only if class $c$ occurs in the batch. The converse does not hold, i.e., a class that occurs can have a positive entry when the other samples assign it a large probability. Both attacks in Section~\ref{sec:privacy} exploit this structure. Throughout this report, we write pp for percentage points, i.e., for absolute differences between two percentages.

\section{Stage 1: Centralized Fine-Tuning}
\label{sec:stage1}

\subsection{Protocol}
The complete network is fine-tuned on the Stage~1 training split (40{,}000 images) with AdamW \citep{loshchilov2019adamw} (learning rate $3\times10^{-4}$, weight decay $10^{-4}$), a cosine learning-rate schedule over 15 epochs \citep{loshchilov2017sgdr}, cross-entropy loss, and batch size 128. After every epoch we evaluate the model on all three splits. The checkpoint with the highest validation accuracy is retained, and training stops early if validation accuracy does not improve for four consecutive epochs. The test split is evaluated after every epoch for monitoring only and is not used for any decision. Training metrics are computed on the augmented training stream and therefore measure performance on randomly cropped and jittered views. Before fine-tuning, we evaluate the pretrained backbone with the untrained 10-class head as a reference.

\subsection{Results}
\begin{figure}[!tbp]
\centering
\includegraphics[width=\linewidth]{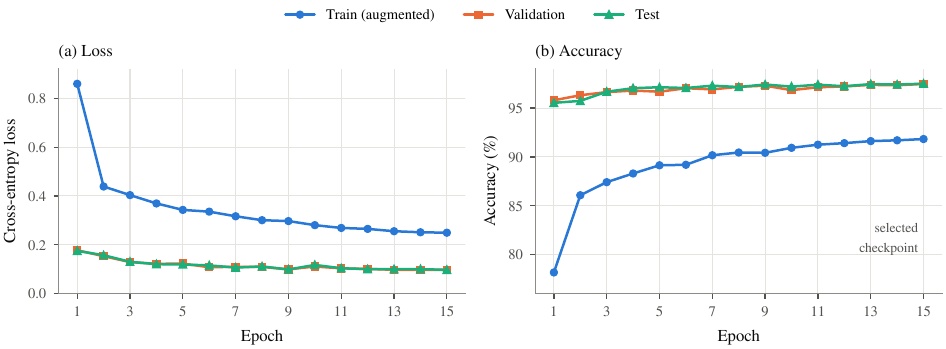}
\caption{Stage 1 learning curves. (a) Cross-entropy loss and (b) accuracy per epoch.}
\label{fig:stage1}
\end{figure}

Table~\ref{tab:stage1} and Figure~\ref{fig:stage1} summarize Stage~1. With the untrained head, the model performs at or below chance level (6.00\% test accuracy against 10\% for uniform guessing), with a cross-entropy of about 8.6, far above the $\ln 10\approx2.30$ of a uniform predictor. The randomly initialized head is therefore confidently wrong rather than uninformative. One epoch of fine-tuning raises validation accuracy to 95.80\%, and accuracy increases slowly thereafter. Validation accuracy reaches its maximum of 97.48\% at the last epoch. As a result, the early-stopping counter never reaches its patience of four (its maximum is three, at epoch 12), and the selected checkpoint is the epoch-15 model. The selected model attains 97.50\% test accuracy and 0.0969 test loss. Validation and test curves agree closely throughout training (largest gap of 0.58\pp, at epoch 2), which indicates that model selection did not overfit the validation split.

Training accuracy stays below validation accuracy (92.29\% against 97.48\% for the selected checkpoint).
The training metric is measured on augmented views, and random resized crops with scale down to 8\% can remove most of the strokes of a sparse sketch. Since accuracy was still increasing when the 15-epoch budget ended, a longer schedule would probably improve the model slightly. We kept the budget fixed so that the training loop completes in under one hour on a single Colab GPU (Section~\ref{sec:resources}). Reloading the exported \texttt{safetensors} checkpoint \citep{safetensors} reproduces the test accuracy exactly. This checkpoint, together with a configuration file that records the class order and preprocessing statistics, constitutes the global initialization $\theta_0$ of all later stages.

\begin{table}[!tbp]
\centering
\caption{Stage 1 performance before fine-tuning (pretrained backbone, untrained head) and after fine-tuning (selected checkpoint, epoch 15).}
\label{tab:stage1}
\small
\begin{tabular}{@{}lccccc@{}}
\toprule
Split & Loss before & Loss after & Accuracy before (\%) & Accuracy after (\%) & $\Delta$ accuracy (pp) \\
\midrule
Train (augmented) & 7.8110 & 0.2378 & 6.97 & 92.29 & $+85.32$ \\
Validation & 8.5807 & 0.0961 & 5.64 & 97.48 & $+91.84$ \\
Test & 8.5695 & 0.0969 & 6.00 & 97.50 & $+91.50$ \\
\bottomrule
\end{tabular}
\end{table}

\section{Stage 2: Clean Federated Fine-Tuning}
\label{sec:stage2}

Stage~2 establishes the attack-free federated baseline against which the poisoning and defense experiments are compared. This stage models a provider that distributes the Stage~1 model to users whose drawings must remain on their devices.

\subsection{Client population}
\begin{figure}[!tbp]
\centering
\includegraphics[width=\linewidth]{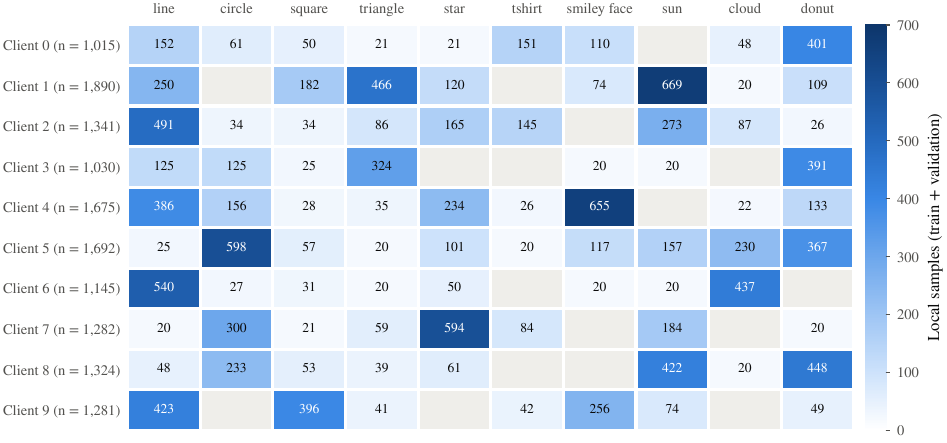}
\caption{Client label distribution of the federated population. Each cell gives the number of local samples (training and validation) of a class, and gray cells mark classes that a client does not hold. Classes per client range from 7 to 10 and dataset sizes from 1{,}015 to 1{,}890 samples.}
\label{fig:clients}
\end{figure}

We simulate $K=10$ clients (\nb{4}). For each client, we draw a dataset size uniformly from $[1{,}000, 2{,}000]$ and a number of classes uniformly from $\{5,\dots,10\}$, and select that many classes at random. We then assign 20 samples to each selected class and distribute the remaining samples according to class proportions drawn from a Dirichlet distribution with concentration $\alpha=0.35$ \citep{hsu2019dirichlet}. The resulting population is heterogeneous both in size (1{,}015 to 1{,}890 samples) and in label composition (Figure~\ref{fig:clients}). For example, \cls{sun} accounts for 35\% of client~1 and \cls{smiley\_face} for 39\% of client~4, whereas several clients hold only the minimum of 20 samples for some classes.

To make the federated stage a meaningful adaptation problem, clients do not receive arbitrary samples. For each class, we draw at least 5{,}000 random candidates with index $\geq10{,}000$ and score them with the cross-entropy loss of the Stage~1 model. We keep the required number of highest-loss candidates and shuffle them before assignment. The retained fraction ranges from 9.4\% of the candidates for \cls{tshirt} (468 of 5{,}000) to 49.2\% for \cls{line} (2{,}460 of 5{,}000). Client data are therefore harder than the Stage~1 test set. With the same preprocessing, the Stage~1 model reaches 90.34\% accuracy on the pooled client training data, compared with 97.66\% on the global evaluation set (Section~\ref{sec:data}). Each client splits its data 90/10 into local training and validation sets, stratified by class, which gives 12{,}310 training and 1{,}365 validation samples in total. 

\subsection{Head-only FedAvg}
The global head is initialized from the Stage~1 checkpoint, $\theta_0=(W_0,b_0)$, and the backbone is frozen. Since a frozen backbone in evaluation mode is a deterministic function, we compute the embeddings of all client and test images once and train the head directly on them. Such a procedure produces the optimization problem of head-only fine-tuning through the frozen network at a fraction of the cost. In every round $t=1,\dots,50$, each client $k$ copies the current global head, trains it for one local epoch with Adam \citep{kingma2015adam} (learning rate $10^{-3}$, batch size 128, no weight decay, optimizer state re-initialized every round), and returns its head $\theta_{t}^{k}$. The server aggregates the heads with sample-weighted FedAvg,
\begin{equation}
    \theta_{t} = \sum_{k=1}^{K}\frac{n_k}{\sum_{j=1}^{K}n_j}\,\theta_{t}^{k},
    \label{eq:fedavg}
\end{equation}
where $n_k$ is the number of local training samples of client $k$. We report the sample-weighted loss and accuracy over all client training and validation splits, which measure adaptation to the federated distribution. We also report the loss and accuracy on the global evaluation set (Section~\ref{sec:data}), which measure whether the federated stage preserves the behavior learned in Stage~1. Since the pooled client validation set contains 1{,}365 images, one image corresponds to 0.073\pp\ of validation accuracy, and on the 5{,}000-image global evaluation set, one image corresponds to 0.02\pp.

\subsection{Results}
\begin{figure}[!tbp]
\centering
\includegraphics[width=\linewidth]{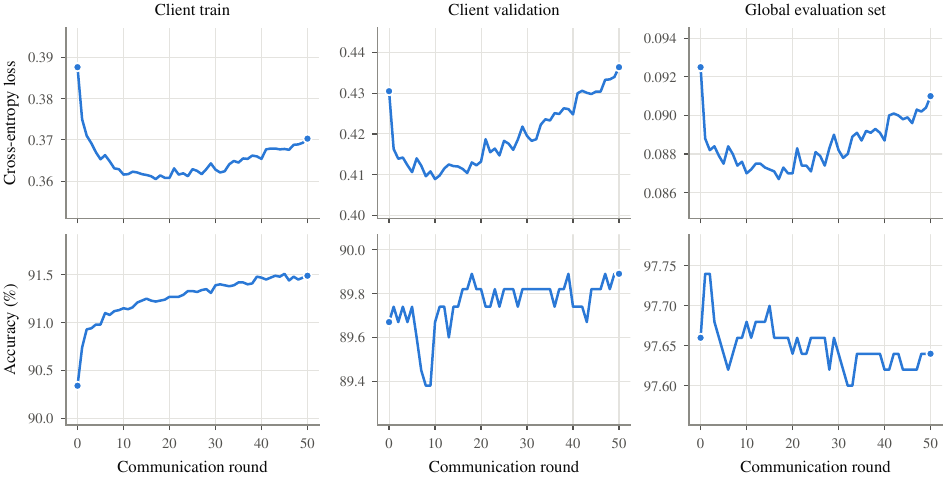}
\caption{Clean federated fine-tuning over 50 rounds. Columns show the pooled client training and validation data, and the global evaluation set, while rows show cross-entropy loss and accuracy.}
\label{fig:clean-fl}
\end{figure}

Table~\ref{tab:clean-fl} and Figure~\ref{fig:clean-fl} report the clean federated run. After 50 rounds, client training accuracy rises from 90.34\% to 91.49\% ($+1.15$\pp) and client training loss falls from 0.3876 to 0.3703. More than half of this gain (0.64\pp) is achieved in the first four rounds. Client validation accuracy changes from 89.67\% to 89.89\%, a difference of three images. Global accuracy stays between 97.60\% and 97.74\% in every round and ends at 97.64\%, i.e., one image below its initial value.

The loss curves show a mild trade-off between personalization and generalization. Client training loss reaches its minimum of 0.3605 at round 17 and then increases slowly, while client validation loss is lowest at round 10 (0.4089) before rising to 0.4364. Accuracy does not decrease over the same rounds, which is consistent with growing confidence on a few misclassified examples. Overall, head-only FedAvg adapts the model modestly to the hard client distribution and preserves its performance on the Stage~1 data distribution. A limit of roughly one percentage point on the training side is plausible for this setting, since only a linear head over fixed features is trained and the selected client data are, by construction, the samples that these features separate least well.

\begin{table}[!tbp]
\centering
\caption{Clean head-only federated fine-tuning before the first round and after 50 FedAvg rounds. Client metrics are sample-weighted over all ten clients.}
\label{tab:clean-fl}
\small
\begin{tabular}{@{}lccccc@{}}
\toprule
& \multicolumn{2}{c}{Round 0} & \multicolumn{2}{c}{Round 50} & \\
\cmidrule(lr){2-3}\cmidrule(lr){4-5}
View & Loss & Accuracy (\%) & Loss & Accuracy (\%) & $\Delta$ accuracy (pp) \\
\midrule
Client train (12{,}310) & 0.3876 & 90.34 & 0.3703 & 91.49 & $+1.15$ \\
Client validation (1{,}365) & 0.4305 & 89.67 & 0.4364 & 89.89 & $+0.22$ \\
Global evaluation (5{,}000) & 0.0925 & 97.66 & 0.0910 & 97.64 & $-0.02$ \\
\bottomrule
\end{tabular}
\end{table}

\section{Privacy: Reconstructing Private Data from Updates}
\label{sec:privacy}

\subsection{Threat model and gradient recovery}
\label{sec:threat-model}
The adversary is an honest-but-curious server, i.e., it executes the protocol faithfully but it can analyse any information it receives. The server knows the architecture and the global model $\theta_t$, the global label space, the learning rate $\eta$ of the clients, and the number of local samples, which FedAvg needs for weighting. It also has access to public QuickDraw data, which is realistic since the dataset and the label space of the deployed model are public. However, it does not observe raw images or labels, and it does not modify the model it sends to clients. 

Both attacks rely on the following observation. If client $k$ performs exactly one plain SGD step with learning rate $\eta$ on its head, its update is $\Delta_k=\theta_t^{k}-\theta_t=-\eta\,g^{k}$, where $g^k$ is the gradient of Eq.~\eqref{eq:head-grad} on its local batch. The server can recover the gradient as:
\begin{equation}
    g^{k} = -\Delta_k/\eta .
    \label{eq:delta-to-grad}
\end{equation}
If $m$ clients in a set $S_t$ each perform such a step and the server applies Eq.~\eqref{eq:fedavg}, the change of the global model is
\begin{equation}
    \theta_{t+1}-\theta_t = \sum_{k\in S_t}\frac{n_k}{N}\Delta_k = -\eta\sum_{k\in S_t}\frac{n_k}{N}g^{k},
    \qquad N=\sum_{k\in S_t}n_k,
    \label{eq:agg-grad}
\end{equation}
which is $-\eta$ times the mean-loss gradient of the pooled virtual batch of all $N$ participating samples. The server can therefore compute this pooled gradient without observing any individual update. In \nb{2}, the client performs an actual SGD step, and we experimentally verify Eq.~\eqref{eq:delta-to-grad}. Specifically, the mean squared difference between the delta-derived gradient and the gradient captured inside the client is $0.0$ for both $W$ and $b$. In \nb{3}, the client updates are computed analytically as $-\eta g^{k}$, so Eq.~\eqref{eq:agg-grad} holds by construction, and the corresponding check (also $0.0$) confirms the implementation of the aggregation. Note that the equivalence is specific to one local SGD step with a shared learning rate. With several local steps, momentum or adaptive optimizers, or heterogeneous learning rates, the update is no longer a scaled gradient. Clients therefore use one SGD step in the privacy experiments, whereas Stage~2 uses one epoch of Adam.

\subsection{Attack 1: a single client update}
\label{sec:attack1}

\begin{figure}[!tbp]
\centering
\includegraphics[width=0.7\linewidth]{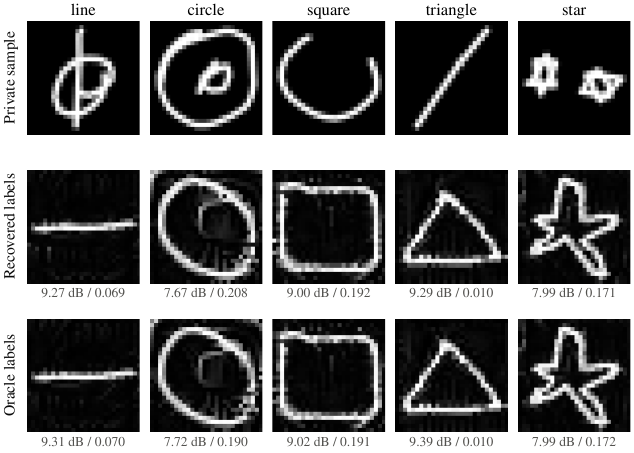}
\caption{Attack 1: private samples (top) and reconstructions from the single-client update using the recovered labels (middle) and the true labels (bottom, evaluation-only diagnostic).}
\label{fig:attack1-recon}
\end{figure}

\paragraph{Setup.}
The target client holds $n=5$ private sketches, one each of \cls{line}, \cls{circle}, \cls{square}, \cls{triangle}, and \cls{star} (\nb{2}). Each sample is the highest-loss bitmap among 128 random candidates of its class, which stresses the attack with informative gradients. Under the Stage~1 model these samples have losses between 1.22 (\cls{star}) and 12.90 (\cls{circle}), and all five are misclassified, as \cls{star}, \cls{donut}, \cls{smiley\_face}, \cls{line}, and \cls{cloud}, respectively. The top row of Figure~\ref{fig:attack1-recon} shows that several of these drawings are atypical for their class, i.e., the \cls{triangle} sample is a single diagonal stroke, the \cls{square} sample is an open arc, and the \cls{line} sample overlays a circle. The client keeps the backbone frozen and the network in evaluation mode, so that batch-normalization statistics are not updated by five samples, and takes one SGD step on the head with $\eta=10^{-2}$. The step reduces the local loss from 5.4683 to 2.8258 and corrects one prediction (\cls{circle}).
\begin{figure}[!tbp]
\centering
\includegraphics[width=0.93\linewidth]{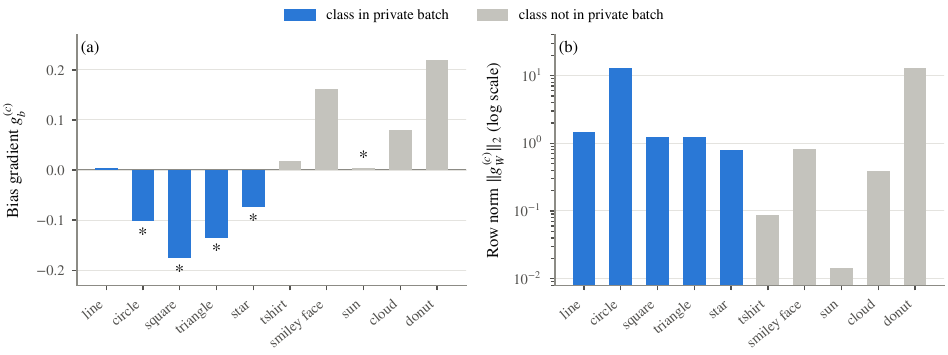}
\caption{Label-leakage signal in a single head update. (a) Bias gradient $g_b^{(c)}$ per class. Asterisks mark the five most negative entries selected by the naive rule, which includes the absent class \cls{sun} and misses \cls{line}. (b) Norm of the weight-gradient row $g_W^{(c)}$ on a logarithmic scale. Both \cls{circle} (present) and \cls{donut} (absent) have large norms because the private \cls{circle} sample is confidently predicted as \cls{donut}.}
\label{fig:attack1-signal}
\end{figure}

\paragraph{Label-set recovery.}
Figure~\ref{fig:attack1-signal} shows the per-class evidence available to the server after applying Eq.~\eqref{eq:delta-to-grad}. The iDLG-based rule of selecting the $n$ most negative bias-gradient entries \citep{zhao2020idlg} returns \cls{square}, \cls{triangle}, \cls{circle}, \cls{star}, and \cls{sun}, i.e., four of the five labels. It misses \cls{line} because $g_b^{(\cls{line})}=+0.0049$ is positive. By Eq.~\eqref{eq:head-grad}, $g_b^{(c)}=\tfrac{1}{n}\big(\sum_i p_i^{(c)}-1\big)$ for a class that occurs once, and the confident prediction of the private \cls{triangle} sample as \cls{line} pushes the summed probability of \cls{line} slightly above one. The absent class \cls{sun} has the smallest positive entry ($+0.0038$) and is selected instead. We therefore score complete label sets. For every candidate set $S$ of size $n$,
\begin{equation}
    \mathrm{score}_\lambda(S) = -\sum_{c\in S}g_b^{(c)} + \lambda\sum_{c\in S}\big\lVert g_W^{(c)}\big\rVert_2 ,
    \label{eq:set-score}
\end{equation}
and the attacker returns the highest-scoring set. With $C=10$ and $n=5$ there are only $\binom{10}{5}=252$ candidate sets, so the search is exhaustive. The first term is the bias evidence. The second term rewards classes whose weight-gradient row is large, which is the case for classes involved in errors. The row norms should be weighted lightly, because they are large for confusions as well as for present classes. For example, the misclassified \cls{circle} sample yields row norms of 12.89 for \cls{circle} and 12.90 for the absent class \cls{donut} (Figure~\ref{fig:attack1-signal}b). With the weight $\lambda=0.01$, the true set ranks first (score 0.6597), ahead of the set that replaces \cls{line} by \cls{sun} (score 0.6464). Table~\ref{tab:attack1-labels} summarizes both rules.

The margin is small, and the result depends on $\lambda$. Recomputing Eq.~\eqref{eq:set-score} from the per-class values shows that the true set is the arg max only for $0.00079\leq\lambda\leq0.0188$. For $\lambda=0$ the score reduces to the naive rule and the true set ranks second. For $\lambda>0.0188$, the large \cls{donut} row norm pulls \cls{donut} into the selected set. Hence, 4 of 5 labels are recovered by the parameter-free rule and the remaining ambiguity concerns the two near-zero entries for \cls{line} and \cls{sun}.

\begin{table}[!tbp]
\centering
\caption{Attack 1: label-set recovery from one single-client update. The margin is the score difference to the second-ranked set.}
\label{tab:attack1-labels}
\small
\begin{tabular}{@{}lllc@{}}
\toprule
Rule & Recovered label set & Correct & Margin \\
\midrule
Top-$n$ most negative $g_b^{(c)}$ ($\equiv$ Eq.~\eqref{eq:set-score} with $\lambda=0$) & \cls{square}, \cls{triangle}, \cls{circle}, \cls{star}, \cls{sun} & 4/5 & 0.0011 \\
Set search, Eq.~\eqref{eq:set-score} with $\lambda=0.01$ & \cls{line}, \cls{circle}, \cls{square}, \cls{triangle}, \cls{star} & 5/5 & 0.0133 \\
\bottomrule
\end{tabular}
\end{table}

\paragraph{Image reconstruction.}
Given the recovered labels $\hat S$, the attacker reconstructs one image per label by gradient matching \citep{zhu2019dlg,geiping2020inverting}. Each image is a learnable $28\times28$ canvas of logits $\zeta$. The pixel intensities $u=\sigma(\zeta)\in[0,1]^{28\times28}$ are bilinearly upsampled to $224\times224$, replicated to three channels, normalized, and passed through the frozen backbone and the global head. Let $(\hat g_W,\hat g_b)$ be the head gradient of Eq.~\eqref{eq:head-grad} induced by the canvases and the recovered labels. The attacker minimizes
\begin{equation}
    \mathcal{L}(\zeta) =
    \frac{\lVert\hat g_W-g_W\rVert_2^2}{\lVert g_W\rVert_2^2}
    + \alpha\frac{\lVert\hat g_b-g_b\rVert_2^2}{\lVert g_b\rVert_2^2}
    + \beta_{\mathrm{TV}}\,\mathrm{TV}(u)
    + \beta_{\mathrm{sp}}\,\overline{u}
    + \beta_{\mathrm{bin}}\,\overline{u(1-u)},
    \label{eq:recon-loss}
\end{equation}
where $\mathrm{TV}(u)$ is the mean absolute difference between vertically and horizontally adjacent pixels \citep{rudin1992tv}, $\overline{u}$ is the mean intensity, which favors the mostly black QuickDraw background, and $\overline{u(1-u)}$ penalizes gray values. We use $\alpha=10$, $\beta_{\mathrm{TV}}=0.01$, and $\beta_{\mathrm{sp}}=\beta_{\mathrm{bin}}=0.001$. Each canvas is initialized from an unrelated public sketch of its recovered class with Gaussian logit noise (standard deviation 0.25) and optimized with Adam (learning rate 0.05) for 2{,}500 steps. The best of five restarts, selected by the final gradient-matching loss, is returned. The public initialization supplies a class prior but no information about the private sample. As a diagnostic that isolates the pixel optimization from label recovery, the same procedure is repeated with the true labels (``oracle'').

We compare reconstructions with the private images by mean squared error (MSE) on the $[0,255]$ scale, peak signal-to-noise ratio (PSNR), and a single-window structural similarity index (SSIM) computed from global image statistics, which is a simplification of the windowed SSIM of \citet{wang2004ssim}. For interpretation, we add two references that we computed post hoc from the $28\times28$ arrays rendered in the notebook figures. Recomputing the metrics of the notebook from these arrays reproduces them to within 0.5\% in MSE, 0.02\,dB in PSNR, and 0.001 in SSIM. The first reference is an empty (all-black) canvas, while the second scores every reconstruction against the private sketches of the other classes (mismatched pairs).

\paragraph{Results.}
The gradient-matching term decreases by almost three orders of magnitude, from 10.86 at initialization to 0.0156 for the best restart, so the reconstructions explain the observed update almost perfectly. Figure~\ref{fig:attack1-recon} and Table~\ref{tab:attack1-recon} show that they are nevertheless poor copies of the private drawings. The reconstructions are clean prototypes of the recovered classes, namely a horizontal line, a closed loop, a rectangle, a triangle, and a five-pointed star. The private \cls{triangle} is a diagonal stroke and the private \cls{square} an open arc, so the reconstructions disclose the label and a canonical shape, but not the client's private drawing. The mean SSIM of 0.130 is higher than for mismatched pairs (0.055), so the reconstructions carry some image-specific structure, for example the outer loop of the private \cls{circle} (SSIM 0.208). By MSE and PSNR, however, every reconstruction is worse than an empty canvas, which reaches 10.41\,dB on average against 8.65\,dB. The oracle diagnostic yields practically the same images and metrics (mean SSIM 0.126). The limiting factor is therefore the information content of a head-only update. The head gradient depends on the input only through the embedding $h$, and many different sketches of a class map to embeddings that produce nearly the same head gradient.

\begin{table}[!tbp]
\centering
\caption{Attack 1: reconstruction quality against the private $28\times28$ sketches ($\downarrow$: lower is better, $\uparrow$: higher is better). Rows marked $\dagger$ are post hoc references (Section~\ref{sec:attack1}).}
\label{tab:attack1-recon}
\small
\begin{tabular}{@{}lcccccc@{}}
\toprule
& \multicolumn{3}{c}{Recovered labels} & \multicolumn{3}{c}{Oracle labels} \\
\cmidrule(lr){2-4}\cmidrule(lr){5-7}
Class & MSE $\downarrow$ & PSNR (dB) $\uparrow$ & SSIM $\uparrow$ & MSE $\downarrow$ & PSNR (dB) $\uparrow$ & SSIM $\uparrow$ \\
\midrule
\cls{line}     & 7690.18  & 9.27 & 0.069 & 7620.23  & 9.31 & 0.070 \\
\cls{circle}   & 11111.11 & 7.67 & 0.208 & 10979.75 & 7.72 & 0.190 \\
\cls{square}   & 8178.44  & 9.00 & 0.192 & 8155.11  & 9.02 & 0.191 \\
\cls{triangle} & 7658.22  & 9.29 & 0.010 & 7478.31  & 9.39 & 0.010 \\
\cls{star}     & 10332.21 & 7.99 & 0.171 & 10327.46 & 7.99 & 0.172 \\
\midrule
Mean           & 8994.03  & 8.65 & 0.130 & 8912.17  & 8.69 & 0.126 \\
\midrule
Empty canvas$^\dagger$ (mean)      & 6495.6  & 10.41 & 0.000 & \multicolumn{3}{c}{--} \\
Mismatched pairs$^\dagger$ (mean) & 10178.2 & 8.24  & 0.055 & \multicolumn{3}{c}{--} \\
\bottomrule
\end{tabular}
\end{table}

\subsection{Attack 2: a FedAvg aggregate update}
\label{sec:attack2}

\paragraph{Setup.}
Beyond attacking a single client, we build a more realistic federated scenario with a pool of ten clients. Each client holds between four and eight private sketches of distinct classes (8, 4, 5, 6, 5, 7, 5, 6, 5, and 5 for clients 0 to 9), chosen with the same hard selection as in Attack~1. A class can therefore occur at most once per client but several times across clients. Two classes of the global label space, \cls{tshirt} and \cls{donut}, are withheld from every client. The attacker does not know which classes are absent and must infer $q_{\cls{tshirt}}=q_{\cls{donut}}=0$ from the update alone. In one round, the first $m\in\{1,2,5,10\}$ clients each take one SGD step on their head ($\eta=10^{-2}$), and the server applies Eq.~\eqref{eq:fedavg}. The participating sets are nested, so the $N=8$, 12, 28, and 56 private samples for increasing $m$ always include those of smaller $m$. The attacker observes $\theta_t$ and $\theta_{t+1}$, $m$, $N$, and $\eta$, and obtains the pooled gradient $g^{\mathrm{agg}}$ from Eq.~\eqref{eq:agg-grad}. Compared with Attack~1, the hidden object is a label-count vector $q\in\mathbb{N}^{C}$ with $\sum_c q_c=N$ and $0\leq q_c\leq m$, and the gradient mixes the samples of several clients.

\paragraph{Label-count recovery.}
The attacker estimates per-class gradient prototypes from public data. For each class $c$, it passes $R=96$ public sketches through the model and averages their single-sample head gradients, $P^{(c)}=(P_W^{(c)},P_b^{(c)})$. It then models the summed private gradient $Ng^{\mathrm{agg}}$ as a non-negative combination of prototypes and solves
\begin{equation}
\begin{aligned}
    \min_{q\in\mathbb{R}^{C}}\quad
    J(q) = \frac{\big\lVert\sum_{c}q_cP_W^{(c)}-Ng_W^{\mathrm{agg}}\big\rVert_2^2}{\lVert Ng_W^{\mathrm{agg}}\rVert_2^2}
    + \gamma\,\frac{\big\lVert\sum_{c}q_cP_b^{(c)}-Ng_b^{\mathrm{agg}}\big\rVert_2^2}{\lVert Ng_b^{\mathrm{agg}}\rVert_2^2}
    \quad\text{s.t.}\quad \sum_{c}q_c=N,\;\; 0\leq q_c\leq m,
\end{aligned}
\label{eq:count-recovery}
\end{equation}
with $\gamma=10$, using sequential least-squares quadratic programming (SLSQP) \citep{kraft1988slsqp,virtanen2020scipy} initialized at $q_c=N/C$ (Algorithm~\ref{alg:count}). The continuous solution is rounded to integers that preserve the total $N$ and the bounds. We measure label-count accuracy as $\sum_c\min(q_c,\hat q_c)/N$, i.e., the fraction of private samples whose label is accounted for. Two references help interpret this metric. First, an uninformed allocation $\hat q_c=N/C$ already scores $\sum_c\min(q_c,N/C)/N$. The resulting value is 80.0\% for $m\in\{1,5,10\}$ and 73.3\% for $m=2$, because the private classes are spread almost uniformly over the eight allowed classes. Second, the trivial prediction $q=0$ gives $J=1+\gamma=11$, so values of $J$ close to 11 indicate that the prototypes explain little of the observed gradient.

\begin{algorithm}[!tbp]
\caption{Attack 2: label-count recovery from a FedAvg aggregate}
\label{alg:count}
\begin{algorithmic}[1]
\Require global heads $\theta_t,\theta_{t+1}$; learning rate $\eta$; number of clients $m$; total samples $N$; public data per class
\State $g^{\mathrm{agg}}\gets-(\theta_{t+1}-\theta_t)/\eta$ \Comment{Eq.~\eqref{eq:agg-grad}}
\For{each class $c=1,\dots,C$}
    \State $P^{(c)}\gets$ mean single-sample head gradient of $R$ public sketches of class $c$ at $\theta_t$
\EndFor
\State $\tilde q\gets$ SLSQP solution of Eq.~\eqref{eq:count-recovery} from $q_c=N/C$
\State $\hat q\gets$ integer rounding of $\tilde q$ preserving $\sum_c\hat q_c=N$ and $0\leq\hat q_c\leq m$
\State \Return label multiset with $\hat q_c$ copies of class $c$
\end{algorithmic}
\end{algorithm}

\begin{figure}[!tbp]
\centering
\includegraphics[width=\linewidth]{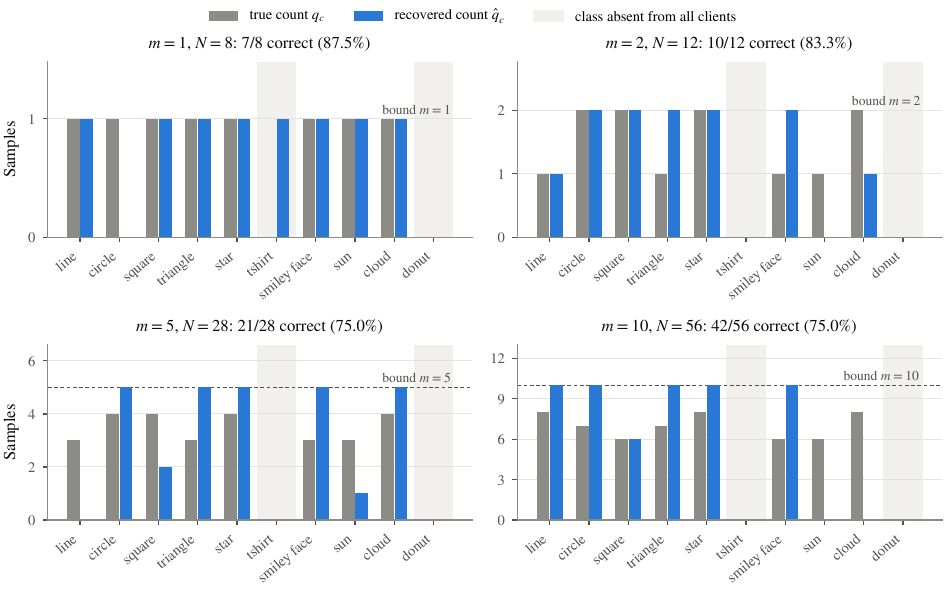}
\caption{Attack 2: true and recovered label counts for $m=1,2,5,10$ participating clients (\nb{3}). Shaded columns mark the classes withheld from all clients. The dashed line is the per-class upper bound $q_c\leq m$. For $m\geq2$, the solution places five classes at the bound and omits classes that occur several times.}
\label{fig:attack2-counts}
\end{figure}

Table~\ref{tab:attack2-counts} and Figure~\ref{fig:attack2-counts} report the results. With a single client the attacker recovers seven of eight labels. It misses the private \cls{circle} sample (an L-shaped stroke) and assigns one count to the withheld class \cls{tshirt}. With more clients, accuracy drops to 83.3\% for $m=2$ and to 75.0\% for $m=5$ and $m=10$. For $m\geq5$ the attack accuracy is below the uninformed uniform allocation, so the estimator does not provide information about the label counts beyond the knowledge of $N$. The solutions have a characteristic structure, i.e., five classes are set to the upper bound $q_c=m$ and the remaining mass is assigned to one or two classes. Classes such as \cls{line} ($m=5$, three occurrences) or \cls{sun} and \cls{cloud} ($m=10$, six and eight occurrences) receive no count. The objective values of 10.20 to 10.84 are only 1.5\% to 7.3\% below the trivial value of 11, which shows that the prototype mixture fits the aggregate gradient poorly.

We attribute the misfit to the hard selection of the private samples. Prototypes average the gradients of typical public sketches, most of which the model classifies correctly and which therefore have small gradients. The private gradients, in contrast, are dominated by a few confidently misclassified samples. Matching their magnitude requires inflating the counts of the classes whose prototypes best align with the aggregate direction until the bound is reached. The attack is nevertheless informative. From $m=2$ onwards it never assigns a count to the withheld classes, and the withheld class \cls{donut} always receives zero, consistent with its large positive bias gradient. Its output therefore reveals which classes are present more reliably than how often they occur.

\begin{table}[!tbp]
\centering
\caption{Attack 2: label-count recovery from a one-round FedAvg aggregate (\nb{3}). ``Uniform'' is the accuracy of the uninformed allocation $\hat q_c=N/C$, and $J(\tilde q)$ is the objective of Eq.~\eqref{eq:count-recovery} at the continuous solution, to be compared with $J=11$ for $q=0$.}
\label{tab:attack2-counts}
\small
\begin{tabular}{@{}cccccccc@{}}
\toprule
$m$ & Clients & $N$ & Correct & Count accuracy (\%) & Uniform (\%) & $J(\tilde q)$ & Classes at bound $m$ \\
\midrule
1  & 0      & 8  & 7/8   & 87.5 & 80.0 & 10.84 & --$^{\ast}$ \\
2  & 0--1   & 12 & 10/12 & 83.3 & 73.3 & 10.45 & 5 \\
5  & 0--4   & 28 & 21/28 & 75.0 & 80.0 & 10.20 & 5 \\
10 & 0--9   & 56 & 42/56 & 75.0 & 80.0 & 10.44 & 5 \\
\bottomrule
\multicolumn{8}{l}{\footnotesize $^{\ast}$For $m=1$, every non-zero count equals the bound.}
\end{tabular}
\end{table}

\paragraph{Image reconstruction.}
The attacker expands $\hat q$ into a label multiset and optimizes one canvas per label with Eq.~\eqref{eq:recon-loss}, using the aggregate gradient as target and the same regularization weights as in Attack~1. Canvases are initialized from public sketches with logit noise of standard deviation 0.20. Since the number of canvases grows with $N$, we reduce the budget with $m$ (2{,}500, 2{,}000, 1{,}500, and 1{,}000 Adam steps with 5, 3, 2, and 1 restarts for $m=1,2,5,10$) and use a learning rate of 0.01 with gradient clipping at norm 1. Instead of the final iterate, we return the canvases with the lowest objective over all steps and restarts. The aggregate does not indicate which reconstruction belongs to which sample, so reconstructions are matched to private samples of the same class by minimum-MSE assignment \citep{kuhn1955hungarian}. We report the metrics over matched pairs and a ``full'' variant, in which every private sample whose class received too few counts is compared with an empty canvas.

\begin{figure}[!tbp]
\centering
\includegraphics[width=0.9\linewidth]{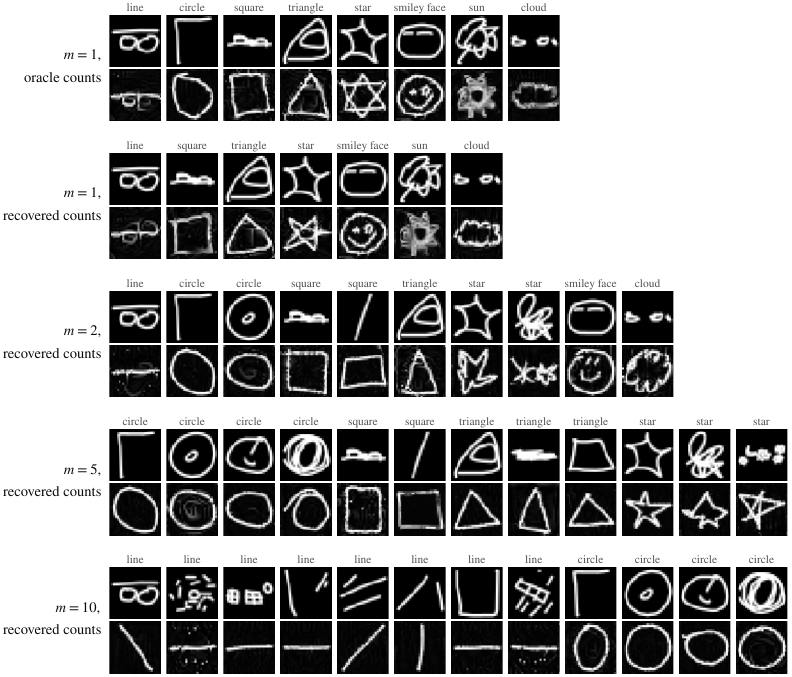}
\caption{Attack 2: private samples (upper row of each block) and reconstructions from the FedAvg aggregate (lower row), matched within each class by minimum MSE. The first block uses the true label counts for $m=1$ (evaluation only), and the others use the recovered counts.}
\label{fig:attack2-recon}
\end{figure}

Table~\ref{tab:attack2-recon} and Figure~\ref{fig:attack2-recon} show the results. As in Attack~1, the reconstructions are recognizable prototypes of the assigned classes (straight lines, ellipses, rectangles, triangles, and stars), while the atypical private drawings are not recovered. This picture hardly changes with $m$. Two aspects of the evaluation explain why the metrics are flat. First, with the class prior doing most of the work, a reconstruction is about as similar to any private sample of its class as to the sample it is matched to, and matching by minimum MSE favors such generic shapes. Second, the empty-canvas penalty is not conservative for MSE and PSNR. For $m=1$, the full metric (8034.18, 9.23\,dB) is better than the matched-only metric (8615.64, 8.81\,dB), because the missed \cls{circle} sample is a sparse stroke whose empty-canvas MSE (3964, derived from the two reported averages) is lower than the MSE of each of the seven matched reconstructions. Only SSIM penalizes the missed samples consistently. A few reconstructions nonetheless show traces of their private sample. For $m=1$, the \cls{line} reconstruction contains the two loops of the private drawing, and the \cls{triangle} reconstruction has a curved side similar to the private sketch. Such details appear with both recovered and oracle counts and become less visible for larger $m$.

\begin{table}[!tbp]
\centering
\caption{Attack 2: reconstruction quality against the private sketches. ``Matched'' averages over reconstructions assigned to a private sample of the same class, while ``Full'' also includes each missed private sample, scored against an empty canvas.}
\label{tab:attack2-recon}
\small
\begin{tabular}{@{}ccccccccc@{}}
\toprule
& & \multicolumn{3}{c}{Matched pairs} & \multicolumn{3}{c}{Full (missed $\to$ empty canvas)} & \\
\cmidrule(lr){3-5}\cmidrule(lr){6-8}
$m$ & Pairs & MSE $\downarrow$ & PSNR (dB) $\uparrow$ & SSIM $\uparrow$ & MSE $\downarrow$ & PSNR (dB) $\uparrow$ & SSIM $\uparrow$ & Runtime (s) \\
\midrule
1  & 7/8   & 8615.64  & 8.81 & 0.212 & 8034.18  & 9.23 & 0.186 & 393.9 \\
2  & 10/12 & 10081.14 & 8.17 & 0.117 & 10285.06 & 8.08 & 0.097 & 86.0 \\
5  & 21/28 & 8087.35  & 9.32 & 0.198 & 8281.28  & 9.18 & 0.148 & 53.1 \\
10 & 42/56 & 8672.74  & 8.99 & 0.219 & 8611.48  & 9.07 & 0.164 & 33.9 \\
\bottomrule
\end{tabular}
\end{table}

\subsection{Summary of the privacy results}
Under the stated assumptions, a single head-only client update leaks its label set almost completely. Such leakage is the most reliable form observed in our experiments and requires little effort, using only the update and the public label space. Pixel reconstruction, in contrast, reproduces the class prior rather than the private drawing, because a linear head observes the input only through a 1280-dimensional embedding that is invariant to much of the stroke-level detail. FedAvg aggregation across as few as five clients reduces label-count recovery to the level of an uninformed guess for the estimator. Note that these conclusions are specific to head-only updates, a single SGD step, and an attacker that does not tamper with the model. Full-model updates, which expose convolutional gradients, and actively malicious servers are known to leak considerably more \citep{geiping2020inverting,yin2021gradinversion,huang2021evaluating}.

\section{Integrity: Data Poisoning under FedAvg}
\label{sec:poisoning}

\subsection{Threat model}
\label{sec:attack3}
We now consider active adversaries. A fraction $f\in\{0,0.1,0.3,0.5,1.0\}$ of the ten Stage~2 clients are considered malicious (\nb{5}). Malicious clients modify their local training data, train the model by following the protocol, and upload the resulting heads, while the server runs the FedAvg protocol of Section~\ref{sec:stage2}. Each value of $f$ is an independent 50-round run that starts from the Stage~1 head. For each $f$, we select the malicious clients with a random generator, and the resulting sets, listed in Table~\ref{tab:poisoning}, are therefore not nested. With $f=0.3$, for example, clients 4, 5, and 8 are malicious, whereas $f=0.5$ selects clients 0, 1, 2, 8, and 9.

\subsection{Attacks}
\begin{figure}[!tbp]
\centering
\includegraphics[width=0.9\linewidth]{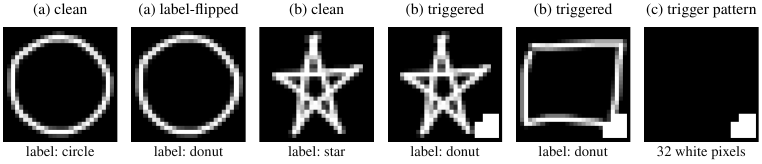}
\caption{Poisoning mechanisms illustrated on Stage~1 sketches. (a) Label flipping changes only the label of \cls{circle} samples to \cls{donut}. (b) Backdoor poisoning adds a fixed trigger to 20\% of the training samples of a malicious client, drawn from its non-\cls{donut} samples, and relabels them as \cls{donut}. (c) A 32-pixel trigger pattern.}
\label{fig:poison-examples}
\end{figure}

\textbf{Label flipping} \citep{tolpegin2020datapoison} relabels every \cls{circle} training sample of a malicious client as \cls{donut}, while the images remain unchanged. \textbf{Backdoor poisoning} \citep{gu2017badnets} selects 20\% of the training samples of a malicious client at random among those whose label is not \cls{donut}, sets a fixed 32-pixel pattern in the bottom-right corner to white, and relabels them as \cls{donut} (Figure~\ref{fig:poison-examples}). The pattern consists of a $4\times4$ square one pixel from the border, extended by two-pixel bands above and to its left. The extensions make it unlikely to coincide with natural strokes. Since the backdoor changes pixels, the embeddings of triggered images are recomputed through the frozen backbone.

Since the backbone is frozen, the attacker can only exploit features that the Stage~1 backbone already computes. A backdoor can succeed only if the trigger moves embeddings in a direction that a linear head can associate with \cls{donut}. Such a setting is harder for the attacker than training from scratch, in which the convolutional layers could learn a dedicated trigger detector.

\subsection{Metrics}
After every round we measure the clean accuracy on the 5{,}000-image global evaluation set and the attack success rate (ASR). For label flipping, ASR is the percentage of the 500 clean \cls{circle} evaluation images predicted as \cls{donut}. For the backdoor, it is the percentage of the 4{,}500 non-\cls{donut} evaluation images that are predicted as \cls{donut} after the trigger is added. Of the 500 \cls{circle} evaluation images, 403 were used to train the Stage~1 model, which may make them harder to flip than unseen images. The ASR at $f=0$ is the background rate at which the model produces the target prediction without an attack. One image changes clean accuracy by 0.02\pp, label-flipping ASR by 0.2\pp, and backdoor ASR by 0.022\pp.
\label{sec:poisoning-results}
\begin{table}[!tbp]
\centering
\caption{Poisoning composition and final results after 50 rounds under FedAvg and FLTrust. ``Weight'' is the share of all client training samples held by malicious clients, i.e., their total FedAvg weight. ``Poisoned'' counts relabeled \cls{circle} samples for label flipping (share of all 1{,}381 \cls{circle} training samples) and triggered samples for the backdoor (share of all 12{,}310 training samples). Acc.\ denotes clean accuracy on the global evaluation set (\%).}
\label{tab:poisoning}
\small
\begin{tabular}{@{}cllrrcccc@{}}
\toprule
& & & & & \multicolumn{2}{c}{FedAvg} & \multicolumn{2}{c}{FLTrust} \\
\cmidrule(lr){6-7}\cmidrule(lr){8-9}
Attack & $f$ & Malicious clients & Weight & Poisoned & Acc. & ASR (\%) & Acc. & ASR (\%) \\
\midrule
\multirow{5}{*}{\shortstack{Label\\flipping}}
& 0.0 & --            & 0.0\%   & 0 (0.0\%)        & 97.62 & 0.60  & 97.66 & 0.60 \\
& 0.1 & 2             & 9.8\%   & 31 (2.2\%)       & 97.64 & 0.60  & 97.64 & 0.60 \\
& 0.3 & 4, 5, 8       & 34.3\%  & 888 (64.3\%)     & 96.48 & 12.60 & 97.64 & 0.80 \\
& 0.5 & 0, 1, 2, 8, 9 & 50.1\%  & 296 (21.4\%)     & 97.68 & 0.60  & 97.66 & 0.60 \\
& 1.0 & all           & 100.0\% & 1{,}381 (100\%)  & 87.92 & 97.80 & 88.02 & 97.80 \\
\midrule
\multirow{5}{*}{Backdoor}
& 0.0 & --            & 0.0\%   & 0 (0.0\%)        & 97.64 & 0.42  & 97.62 & 0.47 \\
& 0.1 & 2             & 9.8\%   & 242 (2.0\%)      & 97.64 & 0.73  & 97.62 & 0.62 \\
& 0.3 & 4, 5, 8       & 34.3\%  & 843 (6.8\%)      & 97.34 & 21.87 & 97.52 & 1.76 \\
& 0.5 & 0, 1, 2, 8, 9 & 50.1\%  & 1{,}234 (10.0\%) & 97.10 & 31.27 & 97.46 & 4.02 \\
& 1.0 & all           & 100.0\% & 2{,}462 (20.0\%) & 96.84 & 54.71 & 96.48 & 50.20 \\
\bottomrule
\end{tabular}
\end{table}
\begin{figure}[!t]
\centering
\includegraphics[width=\linewidth]{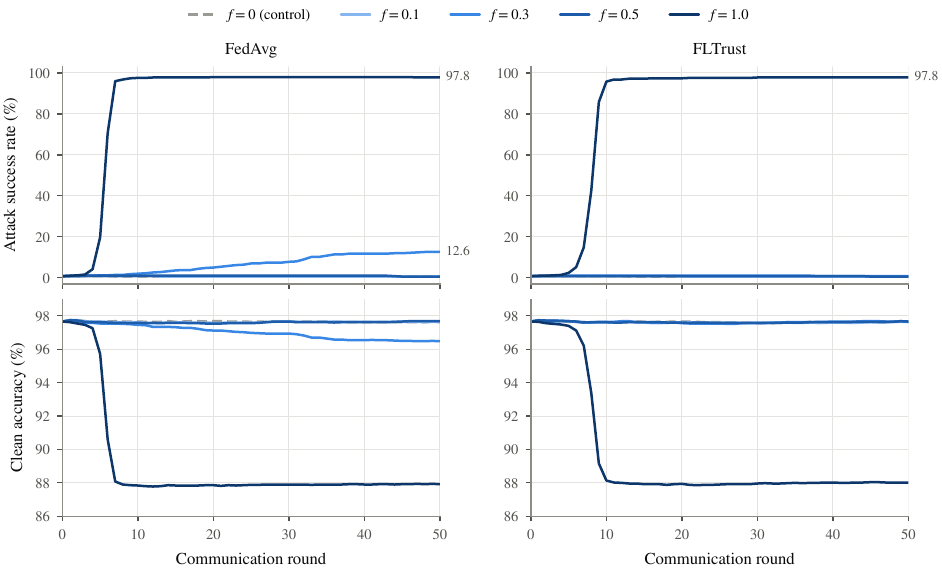}
\caption{Label flipping (\cls{circle}$\to$\cls{donut}) over 50 rounds under FedAvg (left) and FLTrust (right). The top row shows ASR on clean \cls{circle} evaluation images, and the bottom row clean accuracy on the global evaluation set. Color encodes the malicious fraction $f$, and the dashed gray line is the $f=0$ control. End labels give the final ASR where it exceeds 5\%. Under FedAvg, only $f=0.3$ (64\% of \cls{circle} data relabeled) and $f=1.0$ (100\%) have an effect. FLTrust removes the $f=0.3$ effect and delays the $f=1.0$ attack (80\% ASR at round 9 instead of round 7).}
\label{fig:labelflip}
\end{figure}

\subsection{Results}

\begin{figure}[!t]
\centering
\includegraphics[width=\linewidth]{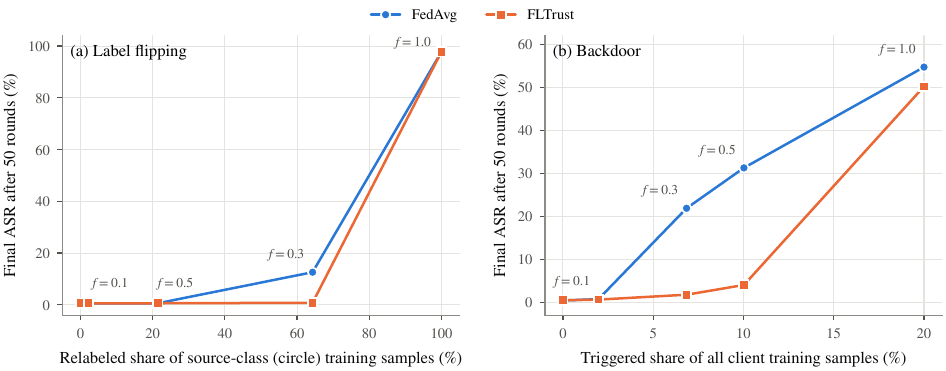}
\caption{Final ASR after 50 rounds as a function of the poisoned data mass. (a) Label flipping against the share of \cls{circle} training samples that are relabeled. The non-nested malicious sets place $f=0.5$ to the left of $f=0.3$. (b) Backdoor against the share of all client training samples that are triggered. Labels give the malicious fraction $f$ of each FedAvg point.}
\label{fig:poison-mass}
\end{figure}

\paragraph{Label flipping.}
Table~\ref{tab:poisoning} (FedAvg columns) and Figure~\ref{fig:labelflip} show that label flipping is ineffective unless it relabels most \cls{circle} data. At $f=0.1$, the single malicious client holds only 31 \cls{circle} training samples, and ASR remains at the background rate of 0.60\% (3 of 500 images). The case $f=0.5$ is equally ineffective, although half of the clients and 50.1\% of the FedAvg weight are malicious. These clients hold only 296 of the 1{,}381 \cls{circle} training samples (21.4\%). In contrast, the three malicious clients at $f=0.3$ hold 64.3\% of the \cls{circle} data and raise ASR to 12.60\% after 50 rounds. ASR is still rising at that point, from 11.80\% at round 40, while clean accuracy falls to 96.48\%. With $f=1.0$, every \cls{circle} sample is relabeled. ASR then exceeds 50\% at round 6 and 80\% at round 7 and stays between 97.80\% and 98.00\% (489 to 490 of 500 images) thereafter, ending at 97.80\%. Clean accuracy drops to 87.92\%, which the attack itself explains. Since \cls{circle} makes up 10\% of the evaluation set, losing 97.8\% of it costs 9.78\pp, close to the observed drop of 9.70\pp\ from the control. Figure~\ref{fig:poison-mass}(a) summarizes this behavior. ASR is governed by the share of source-class training data that is relabeled, and not by the malicious fraction $f$. Such behavior is expected for FedAvg over a frozen backbone since the aggregate approximately minimizes the pooled loss of all clients, and the pooled \cls{circle} data favor \cls{donut} only when most of them carry that label.
\begin{figure}[!t]
\centering
\includegraphics[width=\linewidth]{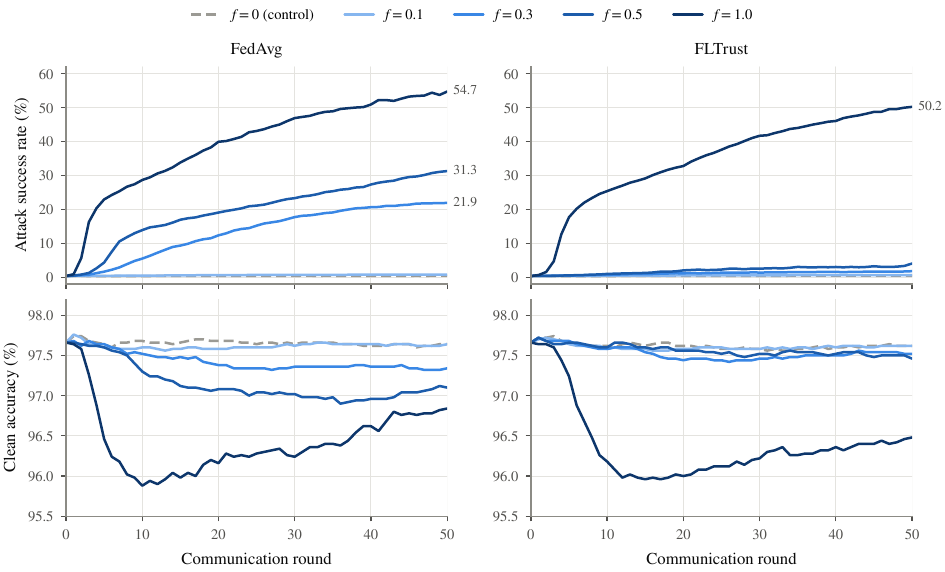}
\caption{Backdoor poisoning over 50 rounds under FedAvg (left) and FLTrust (right). The top row shows ASR on triggered non-\cls{donut} evaluation images, and the bottom row clean accuracy on the global evaluation set. Under FedAvg, ASR grows steadily with $f$, while final clean accuracy stays within 0.8\pp\ of the control (with a transient dip of 1.8\pp\ at $f=1.0$). FLTrust keeps ASR below 4.1\% for $f\leq0.5$, but not for $f=1.0$.}
\label{fig:backdoor}
\end{figure}

\paragraph{Backdoor.}
The backdoor attack behaves differently (Figure~\ref{fig:backdoor}). Its ASR increases monotonically with the amount of triggered data, i.e., 0.73\% at $f=0.1$ (2.0\% of all training samples triggered), 21.87\% at $f=0.3$ (6.8\%), 31.27\% at $f=0.5$ (10.0\%), and 54.71\% at $f=1.0$ (20.0\%), against a background rate of 0.42\%. The final clean accuracy decreases by at most 0.80\pp\ relative to the control (96.84\% at $f=1.0$), and by 0.30\pp\ and 0.54\pp\ at $f=0.3$ and $f=0.5$, which is close to the resolution of a single run. The backdoor is therefore stealthy in the sense that clean accuracy alone would not reveal it. At the same time it remains incomplete. No configuration reaches 80\% ASR within 50 rounds, and the $f=1.0$ run reaches 50\% at round 38. The ASR curves are still increasing at round 50 (for example, from 27.31\% to 31.27\% between rounds 40 and 50 at $f=0.5$), so longer training would likely strengthen the backdoor. The slow growth is consistent with the frozen backbone, since the head must learn the backdoor from the trigger-related variation the Stage~1 features encode. For $f=1.0$, clean accuracy first falls to 95.88\% at round 10 and then partially recovers to 96.84\% as the head learns to separate triggered from clean embeddings.

\section{Defense: FLTrust with a Retained Root Set}
\label{sec:defense}

\subsection{Method}
FLTrust \citep{cao2021fltrust} assumes that the server owns a small clean root dataset and uses it to compute a reference update in every round. In our pipeline this assumption is inexpensive, since the provider performed Stage~1 on clean data and can retain part of it. In \nb{6}, we sample a root set of 100 images per class (1{,}000 in total) from the training part of the Stage~1 split, which is disjoint from the global evaluation set (Table~\ref{tab:provenance}). In each round $t$, the server trains a copy of the current head $\theta_{t-1}$ on the root embeddings with the same local procedure as the clients (one epoch of Adam, learning rate $10^{-3}$, batch size 128) to obtain the trusted update $\Delta_0$. Each client update $\Delta_k=\theta_t^{k}-\theta_{t-1}$ receives the trust score
\begin{equation}
    s_k = \max\!\left(0,\;\frac{\langle\Delta_k,\Delta_0\rangle}{\lVert\Delta_k\rVert_2\,\lVert\Delta_0\rVert_2}\right)
    \label{eq:trust}
\end{equation}
and is rescaled to the norm of the trusted update, $\tilde\Delta_k=(\lVert\Delta_0\rVert_2/\lVert\Delta_k\rVert_2)\,\Delta_k$. The server then updates using
\begin{equation}
    \theta_t = \theta_{t-1} + \frac{\sum_{k=1}^{K}s_k\,\tilde\Delta_k}{\sum_{k=1}^{K}s_k}.
    \label{eq:fltrust}
\end{equation}
If all scores are zero, the server applies $\Delta_0$ instead. Updates that point away from the trusted direction are discarded, and no update can dominate a round through its magnitude. Unlike FedAvg, Eq.~\eqref{eq:fltrust} does not weight clients by sample count, so the influence of a client depends only on the direction of its update. The trusted update itself enters only through the scores and the normalization. Algorithm~\ref{alg:fltrust} summarizes one round. All other settings, including the client datasets, malicious-client sets, poisoning rules, local optimizer, and number of rounds, are identical to Section~\ref{sec:poisoning}.

\begin{algorithm}[!tbp]
\caption{One FLTrust round over classifier heads}
\label{alg:fltrust}
\begin{algorithmic}[1]
\Require global head $\theta_{t-1}$; client heads $\theta_t^{1},\dots,\theta_t^{K}$; root embeddings $\mathcal{D}_0$
\State $\Delta_0\gets\mathrm{LocalAdamEpoch}(\theta_{t-1},\mathcal{D}_0)-\theta_{t-1}$ \Comment{trusted update}
\For{$k=1,\dots,K$}
    \State $\Delta_k\gets\theta_t^{k}-\theta_{t-1}$;\quad $s_k\gets\max(0,\cos(\Delta_k,\Delta_0))$;\quad $\tilde\Delta_k\gets\frac{\lVert\Delta_0\rVert}{\lVert\Delta_k\rVert}\Delta_k$
\EndFor
\If{$\sum_k s_k=0$} \State $\theta_t\gets\theta_{t-1}+\Delta_0$ \Else \State $\theta_t\gets\theta_{t-1}+\sum_k s_k\tilde\Delta_k/\sum_k s_k$ \EndIf
\State \Return $\theta_t$
\end{algorithmic}
\end{algorithm}

\subsection{Results}
\paragraph{Clean utility.}
With no malicious clients, FLTrust reaches 97.66\% (label-flipping control) and 97.62\% (backdoor control) clean accuracy, compared with 97.62\% and 97.64\% under FedAvg (Table~\ref{tab:poisoning}). Replacing sample-weighted averaging by trust-weighted, norm-equalized averaging therefore costs no measurable utility in this setting.

\paragraph{Label flipping.}
FLTrust removes the only partially successful label-flipping attack. At $f=0.3$, ASR stays at 0.80\% instead of 12.60\%, and clean accuracy at 97.64\% instead of 96.48\% (Figure~\ref{fig:labelflip}). The settings that were already ineffective under FedAvg remain at the background rate. When all clients are malicious, the outcome matches FedAvg (97.80\% ASR and 88.02\% clean accuracy). The only difference is timing, since ASR crosses 50\% and 80\% at round 9 instead of rounds 6 and 7.

\paragraph{Backdoor.}
The effect of the defense is most visible for the backdoor (Figure~\ref{fig:backdoor}). ASR is reduced from 21.87\% to 1.76\% at $f=0.3$ and from 31.27\% to 4.02\% at $f=0.5$, while clean accuracy stays within a 0.2\pp\ margin of the attack-free control. Note, however, that ASR under FLTrust increases slowly (from 3.00\% at round 40 to 4.02\% at round 50 for $f=0.5$). The defense therefore slows the backdoor substantially rather than eliminating it. At $f=1.0$, ASR reaches 50.20\%, compared with 54.71\% under FedAvg, and clean accuracy 96.48\%. Figure~\ref{fig:poison-mass} presents both aggregators. Under FLTrust, ASR stays near zero until all clients are malicious.

\begin{figure}[!tbp]
\centering
\includegraphics[width=\linewidth]{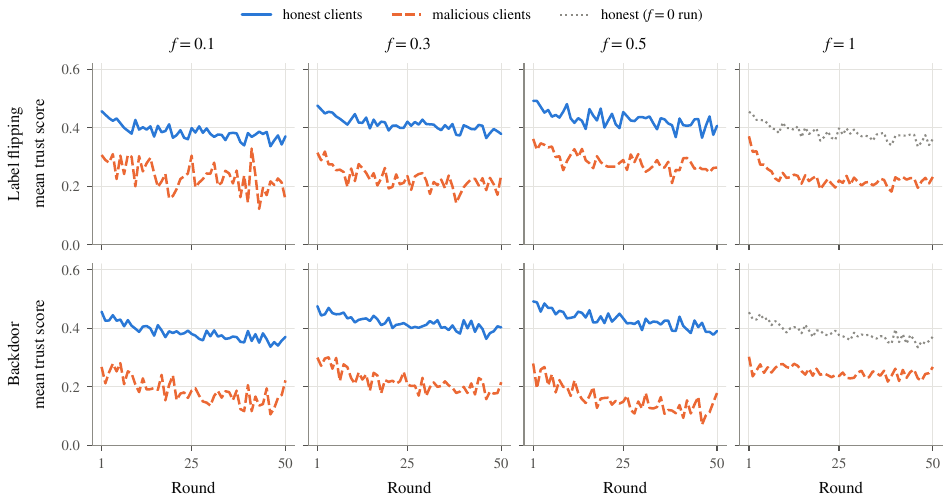}
\caption{FLTrust trust scores (Eq.~\eqref{eq:trust}) per round, averaged over honest (solid) and malicious (dashed) clients, for label flipping (top) and the backdoor (bottom). At $f=1.0$ there are no honest clients, and the dotted line shows the honest trust of the $f=0$ run for reference.}
\label{fig:trust}
\end{figure}

\paragraph{Trust analysis.}
Table~\ref{tab:trust} and Figure~\ref{fig:trust} explain the mechanism. In every run and round, all ten clients received a positive trust score, so FLTrust never excluded a client outright. Malicious updates are nevertheless consistently less aligned with the trusted update. Averaged over rounds, honest clients obtain trust scores of 0.38 to 0.43, and malicious clients 0.16 to 0.29. Since Eq.~\eqref{eq:fltrust} weights norm-equalized updates by trust, malicious clients receive a smaller share of the aggregate than under FedAvg. For the backdoor, their share falls from 34.3\% to 18.2\% at $f=0.3$ and from 50.1\% to 26.6\% at $f=0.5$. The reduction in ASR is larger than the reduction in weight, which indicates that norm equalization also matters. Under FedAvg, the update of a malicious client enters with its full magnitude, whereas under FLTrust it cannot exceed the magnitude of the trusted update. For label flipping at $f=0.5$, malicious trust (0.285) is the highest among the attacked settings, because only 21.4\% of the \cls{circle} data are relabeled and the malicious updates are mostly benign. At $f=1.0$, every update is poisoned, and FLTrust can only reweight poisoned updates among themselves. Their trust remains positive (0.23 to 0.25) because most of the data of each malicious client are clean (80\% for the backdoor and at least 64\% for label flipping). The poisoned updates therefore still agree with the trusted update in most directions. The poisoned component passes into the aggregation, and only its magnitude is limited, which accounts for the small delay of the label-flipping attack and the modest ASR reduction of the backdoor.

\section{Computational Footprint}
\label{sec:resources}

All six notebooks were executed on Google Colab with a GPU runtime, and Table~\ref{tab:resources} lists the runtime and peak memory they recorded. The peak GPU memory allocated by PyTorch remains below 600\,MiB in every notebook, and the federated experiments operate on precomputed embeddings. Any participant can therefore reproduce the complete pipeline in a browser, without a local installation or a dedicated GPU.

Centralized fine-tuning of the full network dominates the cost of the tutorial, at 227.7\,s per epoch on average and 56.9\,min for 15 epochs. On the other hand, the operations that happen on the client side are cheap. A single head-only SGD step on five images takes 0.086\,s and increases allocated GPU memory by 15.3\,MiB. Fifty rounds of head-only FedAvg over ten clients take 12.55\,s for local training and aggregation, because the embeddings are precomputed once. The uploaded head of 12{,}810 parameters (50\,KiB in 32-bit floating point) is 76 times smaller than the full model. FLTrust adds the training of the trusted head to every round, which increases the time of a 50-round run from 11.57\,s to 12.31\,s on average over the ten scenarios (+6.4\%). The privacy attacks are the most expensive operations of the federated stage, because every optimization step back-propagates through the frozen backbone. Their runtime decreases with $m$ only because the optimization budget is reduced, while peak memory grows with the number of canvases.

\begin{table}[!tbp]
\centering
\caption{FLTrust trust diagnostics averaged over 50 rounds. ``FedAvg weight'' is the aggregated weight of malicious clients under Eq.~\eqref{eq:fedavg}. ``FLTrust share'' is their share of the total trust, $\sum_{k\in\mathcal{M}}s_k/\sum_k s_k$, computed per round from the mean honest and malicious trust and averaged over rounds.}
\label{tab:trust}
\small
\begin{tabular}{@{}lccccccc@{}}
\toprule
& & \multicolumn{2}{c}{Mean trust score} & \multicolumn{2}{c}{Malicious aggregation weight} & & \\
\cmidrule(lr){3-4}\cmidrule(lr){5-6}
Attack & $f$ & Honest & Malicious & FedAvg weight & FLTrust share & ASR FedAvg (\%) & ASR FLTrust (\%) \\
\midrule
\multirow{5}{*}{\shortstack{Label\\flipping}}
& 0.0 & 0.384 & --    & 0.0\%   & 0.0\%   & 0.60  & 0.60 \\
& 0.1 & 0.387 & 0.234 & 9.8\%   & 6.3\%   & 0.60  & 0.60 \\
& 0.3 & 0.413 & 0.227 & 34.3\%  & 19.0\%  & 12.60 & 0.80 \\
& 0.5 & 0.429 & 0.285 & 50.1\%  & 39.8\%  & 0.60  & 0.60 \\
& 1.0 & --    & 0.230 & 100.0\% & 100.0\% & 97.80 & 97.80 \\
\midrule
\multirow{5}{*}{Backdoor}
& 0.0 & 0.386 & --    & 0.0\%   & 0.0\%   & 0.42  & 0.47 \\
& 0.1 & 0.385 & 0.187 & 9.8\%   & 5.1\%   & 0.73  & 0.62 \\
& 0.3 & 0.419 & 0.219 & 34.3\%  & 18.2\%  & 21.87 & 1.76 \\
& 0.5 & 0.429 & 0.159 & 50.1\%  & 26.6\%  & 31.27 & 4.02 \\
& 1.0 & --    & 0.250 & 100.0\% & 100.0\% & 54.71 & 50.20 \\
\bottomrule
\end{tabular}
\end{table}

\begin{table}[!tbp]
\centering
\caption{Runtime and peak memory as recorded by the notebooks on a Google Colab GPU runtime. RAM is the peak resident set size of the Python process, and VRAM is the peak memory allocated by PyTorch. Federated runtimes exclude data loading, hard-example selection, and the one-time embedding precomputation. n/r: not recorded.}
\label{tab:resources}
\small
\begin{tabularx}{\linewidth}{@{}l X r r r@{}}
\toprule
Notebook & Operation & Runtime & Peak RAM (MiB) & Peak VRAM (MiB) \\
\midrule
NB1 & Centralized fine-tuning, 15 epochs of 40{,}000 images & 56.9\,min & n/r & n/r \\
NB2 & One head-only SGD step on 5 images (client side) & 0.086\,s & 1288.1 & 35.0 \\
NB3 & Attack 2, complete experiment, $m=1$ / 2 / 5 / 10 & 393.9 / 86.0 / 53.1 / 33.9\,s & 1689--1782 & 100--571 \\
NB4 & Clean FedAvg, 50 rounds, 10 clients (training and aggregation) & 12.55\,s & 2333.5 & 31.3 \\
NB5 & FedAvg under poisoning, per 50-round scenario & 11.19--12.10\,s & $\leq$2444.7 & 31.3 \\
NB6 & FLTrust under poisoning, per 50-round scenario & 12.03--12.70\,s & $\leq$2448.9 & 31.3 \\
\bottomrule
\end{tabularx}
\end{table}

\section{Live Session: Federated Learning with FedMaestro}
\label{sec:fedmaestro-live}

The simulations above provide the controlled and reproducible evidence. The live session complements them by exposing the participants of the summer school to the operational side of FL through FedMaestro, the federated orchestrator of Indigma, which manages model distribution, client participation, round execution, update collection, aggregation, and monitoring. Each participant joins the federated network as a client from their own device, and the session starts from the Stage~1 checkpoint of Section~\ref{sec:stage1}, which serves as the initial global model. The session follows the order of this report, i.e., inference with the centrally trained model, federated fine-tuning, and poisoning. 

\paragraph{Scope.}
The live session is a demonstration of an online federated environment. The number of participants, their drawing styles, and their devices are not controlled, and several of these quantities are smaller than in the simulations (e.g., 5 drawings per client). We therefore report no quantitative results from the session. The statements below describe what the earlier sections lead us to expect, and the reproducible evidence for each of them is the corresponding notebook.

\begin{figure}[t]
\centering
\includegraphics[width=\linewidth]{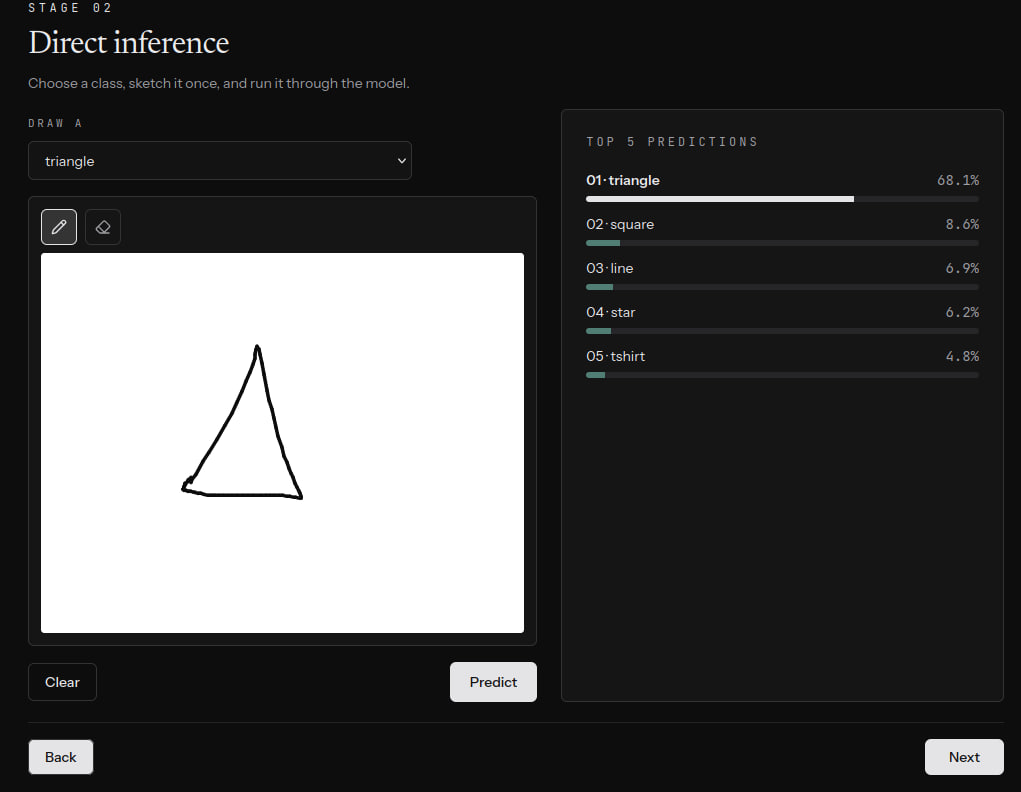}
\caption{Participant view of the live session. The participant selects a class, draws it on the canvas, and receives the five most probable classes under the current model served by the orchestrator, together with their confidences.}
\label{fig:live-inference}
\end{figure}

\paragraph{Step 1: direct inference.}
Participants select one of the ten classes, draw it once on a canvas connected to the deployed Stage~1 model, and receive the five most probable classes with their confidences (Figure~\ref{fig:live-inference}). The drawings are made in black on a white canvas. They are converted to the QuickDraw format, i.e., $28\times28$ grayscale bitmaps with white strokes on a black background, and then processed as in Section~\ref{sec:model}. Even so, they differ from the crowd-sourced training data. They are drawn with a different input device and under no time pressure, and their stroke widths and positions on the canvas follow the participant rather than the original game. We therefore expect drawings of most classes to be recognized. Errors should concentrate on the visually related pairs identified in Section~\ref{sec:data}, such as \cls{circle}/\cls{donut} and \cls{sun}/\cls{star}, and on drawings that depart from the typical shape of a class. Such drawings are also the samples that the Stage~1 model misclassified in Sections~\ref{sec:attack1} and~\ref{sec:attack2}. The ranked list additionally shows which classes the model considers as alternatives, so participants can see model confusions even when the top prediction is correct. This step lets participants identify the classes that the model handles poorly for their own drawings.

\begin{figure}[t]
\centering
\includegraphics[width=\linewidth]{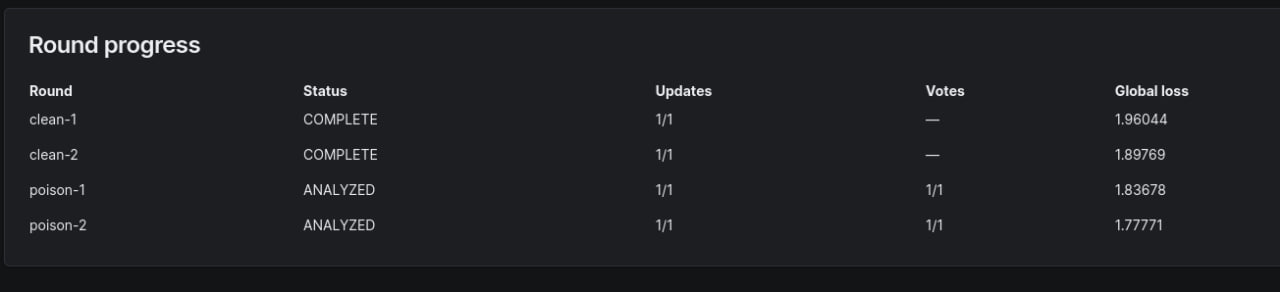}
\caption{Round progress in the FedMaestro administrator dashboard. Clean rounds (Step~2) and poisoning rounds (Step~4) are listed separately with their status, the number of received updates, and the global loss reported by the orchestrator. The screenshot was taken during a test run with a single client.}
\label{fig:fedmaestro-dashboard}
\end{figure}

\paragraph{Step 2: federated fine-tuning on participant drawings.}
Participants contribute new labeled drawings, which remain on their devices and serve as local training data. In every round, FedMaestro distributes the current global model, each participating client trains locally and returns its update, and the server aggregates the updates into a new model version. The administrator dashboard lists every round with its status and the number of updates received from the participating clients (Figure~\ref{fig:fedmaestro-dashboard}). Two points from the simulations are made explicit to participants. First, the model starts from an accurate checkpoint, and fine-tuning adapts it. In Section~\ref{sec:stage2}, head-only FedAvg changed accuracy by about one percentage point over 50 rounds. With far fewer rounds and drawings in the live session, we expect visible changes to be limited to the classes and drawing styles that participants actually contribute. Second, keeping the drawings on the device does not make the update uninformative. Section~\ref{sec:attack1} showed that a single head update reveals which classes its sender drew, while pixel reconstruction recovers only class prototypes. Aggregation over several participants weakens this label leakage (Section~\ref{sec:attack2}).

\paragraph{Step 3: observing whether the model learns.}
After one or more rounds, participants query the updated global model and compare its predictions with those of Step~1. This step connects the abstract FedAvg update of Eq.~\eqref{eq:fedavg} with an observable change in model behavior. The clearest comparison resubmits the drawings of Step~1 to the updated model. We expect the most visible changes on drawings that resemble the contributed training data and on classes that were confused in Step~1. Predictions on canonical drawings should remain stable, consistent with the preserved global accuracy in Section~\ref{sec:stage2}.

\paragraph{Step 4: participant-driven poisoning.}
Finally, a group of participants deliberately poisons its local data, for example by consistently labeling one class as another, which corresponds to the label-flipping attack of Section~\ref{sec:poisoning}. The poisoning rounds run in the same federated network and appear on the administrator dashboard separately from the clean rounds (Figure~\ref{fig:fedmaestro-dashboard}). Participants then check whether the global model starts to produce the target class of the attacker. The simulations lead to two expectations. First, under FedAvg the outcome depends on the share of the source-class data that is relabeled (Figure~\ref{fig:poison-mass}a). A single poisoner should have little effect as long as honest participants contribute drawings of the same class, whereas coordinated relabeling of most source-class drawings should flip the class within a few rounds. Second, the clean Stage~1 data provide the root set that FLTrust requires. With FLTrust, partial attacks are expected to be suppressed, and the influence of a large contribution is bounded by norm equalization (Section~\ref{sec:defense}). An attack in which every participant relabels the same class, however, is expected to succeed under either aggregation rule. Running the poisoning step once with each aggregation rule lets participants compare these outcomes directly.

\section{Discussion}
\label{sec:discussion}

\paragraph{Lessons for post-training federated personalization.}
Three observations carry over beyond this specific pipeline. First, privacy leakage and integrity risk should be assessed for the update that is actually exchanged. A head-only update of 12{,}810 parameters reveals the label composition of a small client batch almost exactly but carries little pixel-level information. In this regime, label privacy rather than image privacy is therefore the primary concern. Second, the strength of a data poisoning attack is determined by how much poisoned data reaches the aggregation and how they are distributed across clients. In non-IID populations, a minority of clients that holds most of a class can cause more damage than a majority that holds little of it (Table~\ref{tab:poisoning}). Third, the centralized stage that precedes federated personalization naturally provides the clean root data that FLTrust requires. In our experiments, the defense suppressed all partial attacks without cost to clean utility.

\paragraph{From convergence to production-ready FL.}
Many FL studies evaluate a federated method mainly by whether and how fast it converges. Our results suggest that a deployment cannot be judged this way alone. The same head-only FedAvg procedure that converges cleanly in Section~\ref{sec:stage2} reveals the label set of a single client to an honest-but-curious server (Section~\ref{sec:attack1}) and accepts a backdoor that clean accuracy alone does not reveal (Section~\ref{sec:poisoning}). In real deployments, curious servers and malicious or faulty clients have to be expected. A production-ready FL system therefore needs components beyond the training loop, such as robust or trust-based aggregation \citep{blanchard2017krum,yin2018byzantine,cao2021fltrust}, filtering and validation of incoming updates, secure aggregation and differential privacy \citep{bonawitz2017secagg,abadi2016dpsgd}, client authentication, and monitoring of every round with the ability to roll back a model version. Each of these components has its own requirements and costs. FLTrust, for example, needs a clean root set and offers no protection when every client is malicious (Section~\ref{sec:defense}). Filters that discard outlying updates can also exclude honest clients whose data are unusual but legitimate, which is a known difficulty in non-IID populations. FL is therefore a complete system rather than an optimization algorithm, and building, operating, and auditing such a system requires engineering effort and expertise beyond those of centralized training \citep{kairouz2021advances,bonawitz2019towards}. The live session of Section~\ref{sec:fedmaestro-live} gives participants a view of this operational side, in which rounds, participating clients, and received updates are tracked explicitly. We believe that this gap between a converging algorithm and a deployable system, more than the learning algorithms themselves, is one reason why many organizations that could benefit from FL do not, or cannot, adopt it, at least for now \citep{perifanis2025bridging}.

\paragraph{Limitations.}
The global evaluation set of \nb{4}--\nb{6} was rebuilt with a different seed than the Stage~1 split and consists mostly of Stage~1 training images (Section~\ref{sec:data}). Global accuracies in the federated experiments are therefore comparable with one another but not with the Stage~1 test accuracy, and rebuilding the split with the Stage~1 seed would remove this caveat. Every configuration was executed once with a fixed seed, so the results carry no confidence intervals. The private samples of the privacy attacks and the client data were selected for high loss under the Stage~1 model. Such a selection strengthens the gradients available to the attacker and makes the federated stage non-trivial, but it also makes the private sketches atypical, which affects the reconstruction metrics. The privacy attacks assume one plain SGD step so that Eqs.~\eqref{eq:delta-to-grad} and~\eqref{eq:agg-grad} hold exactly, whereas Stage~2 uses one epoch of Adam, for which the update is not a scaled gradient. The differentiable rendering of the attacker upsamples canvases directly to $224\times224$ and omits the resize-and-center-crop of the client pipeline. The resulting scale mismatch of about 14\% may weaken pixel reconstruction. The SSIM used throughout is a single-window approximation, and MSE and PSNR favor empty canvases on sparse sketches (Section~\ref{sec:attack1}). The FLTrust root set is drawn from the same distribution as the global evaluation set, which favors the defense. A root set that differs from the client distribution would yield noisier trust scores. Finally, the attackers are not adaptive, i.e., they neither scale their updates nor optimize them to evade a known aggregation rule, and such attacks are known to defeat several robust aggregators \citep{fang2020local,shejwalkar2021manipulating}.

\section{Conclusions and Future Work}
\label{sec:conclusion}

In this report we presented an end-to-end tutorial pipeline in which a pretrained MobileNetV4 is fine-tuned centrally on QuickDraw-10 (97.50\% test accuracy) and then personalized with head-only FedAvg over ten non-IID clients without loss of global accuracy. On this pipeline, a curious server recovered the private label set of a single client update, while pixel reconstructions reproduced class prototypes rather than the private drawings. The aggregation over five or more clients reduced label-count recovery to the level of an uninformed guess. Malicious clients implanted a backdoor that reached 54.7\% ASR at a final clean-accuracy cost below one percentage point, whereas label flipping succeeded only when most of the source-class data were relabeled. FLTrust with a root set retained from centralized fine-tuning suppressed every partial attack without measurable cost, but it could not protect a federation in which every client is malicious. Together with the live FedMaestro session, the six notebooks are intended to make these privacy and integrity risks concrete for the participants of the summer school and to provide a reproducible starting point for studying defenses.

Several open problems remain. The results are based on single runs, on head-only updates, and on attackers that neither tamper with the model nor adapt to the aggregation rule. Future work will extend the tutorial in five directions. First, we plan to add differentially private local training and secure aggregation against the attacks of Section~\ref{sec:privacy}, with an explicit privacy-utility trade-off. Second, we will study partial backbone fine-tuning, which exposes convolutional gradients and allows trigger-specific features. Third, we will compare FLTrust with other robust aggregation mechanisms on the same client population. Fourth, we will consider adaptive attackers that constrain their updates to resemble honest ones. Finally, we will repeat all experiments with several seeds and client populations to obtain confidence intervals.

\section*{Code and Data Availability}
The six executed notebooks, including all outputs from which the results in this report are taken, are available at \url{\repourl} under the Apache License~2.0. All notebooks run on Google Colab, so anyone can reproduce the experiments in a browser without a local installation or high-end hardware (Section~\ref{sec:resources}). The QuickDraw data are downloaded from the public Quick, Draw! repository \citep{quickdraw} and remain subject to their CC BY 4.0 license, while the pretrained weights are subject to the license of their distributor \citep{rw2019timm}.

\bibliographystyle{unsrtnat}
\bibliography{references}

\end{document}